\documentclass[11pt]{article}

\usepackage[final]{acl}

\usepackage[utf8]{inputenc}     % UTF-8 encoding
\usepackage[T1]{fontenc}        % Proper font encoding
\usepackage{microtype}          % Better typography
\usepackage{times}              % Standard ACL font
\usepackage{latexsym}
\usepackage{amsmath, amsfonts, amssymb, bbm}
\usepackage{float}
\usepackage{graphicx}
\usepackage{enumitem}
\usepackage{afterpage}
\usepackage{multirow}
\usepackage{inconsolata}        % Nicer monospace font
\usepackage{todonotes}
\usepackage{placeins}
\usepackage{tikz}
\usetikzlibrary{shapes.geometric, shapes.misc, arrows.meta, positioning, fit, backgrounds, calc}

\tikzset{
    basicbox/.style={rectangle, draw=black, thick, rounded corners, align=center, minimum height=0.8cm, font=\small},
    audio/.style={basicbox, fill=gray!20, shape=chamfered rectangle},
    process/.style={basicbox, fill=blue!10},
    judge/.style={basicbox, shape=trapezium, trapezium angle=75, fill=purple!10},
    data/.style={basicbox, dashed, fill=yellow!10},
    bad/.style={basicbox, fill=red!20, draw=red!80!black},
    good/.style={basicbox, fill=green!20, draw=green!80!black},
    arrow/.style={->, >=LaTeX, thick},
    labeltext/.style={font=\footnotesize\itshape, align=center}
}
\usepackage{booktabs, tabularx, array}
\usepackage{calc}               % For \dimexpr calculations

\usepackage{xcolor}
\usepackage{listings}

\definecolor{codegreen}{rgb}{0,0.6,0}
\definecolor{codegray}{rgb}{0.5,0.5,0.5}
\definecolor{codepurple}{rgb}{0.58,0,0.82}
\definecolor{backcolour}{rgb}{0.95,0.95,0.95}
\usepackage{xcolor}
\definecolor{goodgreen}{HTML}{008B45}
\definecolor{badred}{HTML}{C00000}
\definecolor{softblue}{HTML}{E6F0FF}
\definecolor{lightgray}{gray}{0.97}

\newcommand{\lblcol}[2]{\colorbox{#1!15}{\strut\footnotesize\texttt{#2}}}
\definecolor{goodgreen}{HTML}{007A3D}
\definecolor{badred}{HTML}{C9302C}
\definecolor{softblue}{HTML}{F2F6FF}

\lstdefinestyle{jsonstyle}{
  backgroundcolor=\color{backcolour},
  commentstyle=\color{codegreen},
  keywordstyle=\color{magenta},
  numberstyle=\tiny\color{codegray},
  stringstyle=\color{codepurple},
  basicstyle=\ttfamily\footnotesize,
  breaklines=true,
  captionpos=b,
  keepspaces=true,
  numbers=left,
  numbersep=5pt,
  showspaces=false,
  showstringspaces=false,
  showtabs=false,
  tabsize=2
}
\usepackage{algorithm}
\usepackage{algpseudocode}

\usepackage[most]{tcolorbox}

\tcbset{colback=gray!1, colframe=black!10, boxrule=0.3pt, arc=1pt}

\newtcolorbox{claimbox}[1][]{
  colback=gray!5,
  colframe=gray!75,
  fonttitle=\bfseries,
  coltitle=black,
  title={#1},
  boxrule=0.5pt,
  arc=4pt,
  boxsep=5pt,
  left=5pt, right=5pt, top=2pt, bottom=5pt
}

\usepackage{mdframed}

\title{Hearing Between the Lines: Unlocking the Reasoning Power of LLMs for Speech Evaluation}

\author{
  \textbf{Arjun Chandra}\textsuperscript{1*},
  \textbf{Kevin Miller}\textsuperscript{1*},
  \textbf{Venkatesh Ravichandran}\textsuperscript{2}\\
  \textbf{Constantinos Papayiannis}\textsuperscript{2},
  \textbf{Venkatesh Saligrama}\textsuperscript{1}
  \\
  \textsuperscript{1}Boston University,
  \textsuperscript{2}Amazon AGI \\
}

\date{\today}

\begin{document}
\maketitle
\begingroup
\renewcommand\thefootnote{\*}
\footnotetext{
*Equal contribution. Correspondence to \texttt{srv@bu.edu}.\\
\hspace*{\parindent}Code and Data: \href{https://github.com/arjunchandra2/TRACE}{github.com/arjunchandra2/TRACE}
}
\endgroup

\begin{abstract}
Large Language Model (LLM) judges exhibit strong reasoning capabilities but are limited to textual content. This leaves current automatic Speech-to-Speech (S2S) evaluation methods reliant on opaque and expensive Audio Language Models (ALMs). In this work, we propose \textbf{TRACE} (\textbf{T}extual \textbf{R}easoning over \textbf{A}udio \textbf{C}ues for \textbf{E}valuation), a novel framework that enables LLM judges to reason over audio cues to achieve cost-efficient and human-aligned S2S evaluation. To demonstrate the strength of the framework, we first introduce a Human Chain-of-Thought (HCoT) annotation protocol to improve the diagnostic capability of existing judge benchmarks by separating evaluation into explicit dimensions: \emph{content} (C), \emph{voice quality} (VQ), and \emph{paralinguistics} (P). Using this data, TRACE constructs a textual blueprint of inexpensive audio signals and prompts an LLM to render dimension-wise judgments, fusing them into an overall rating via a deterministic policy. TRACE achieves higher agreement with human raters than ALMs and transcript-only LLM judges while being significantly more cost-effective. We will release the HCoT annotations and the TRACE framework to enable scalable and human-aligned S2S evaluation.
\end{abstract}

\section{Introduction}

There has been rapid progress in speech-to-speech (S2S) models in recent years \cite{défossez2024moshispeechtextfoundationmodel, fang2025llamaomniseamlessspeechinteraction, zhan2025anygptunifiedmultimodalllm, zeng2024glm4voiceintelligenthumanlikeendtoend}, offering a natural interface for spoken communication with artificial assistants. However, the current automatic S2S evaluation paradigms suffer from critical drawbacks. Large Language Model (LLM) judges used in S2S evaluation operate on transcripts alone \cite{chen2024voicebenchbenchmarkingllmbasedvoice, liu2025vocalbenchbenchmarkingvocalconversational, hou2025sovabenchbenchmarkingspeechconversation}, making them blind to crucial non-linguistic speech cues such as sarcasm and emotion. To address this, several recent works leverage Audio Language Models (ALMs) \cite{manakul2025audiojudgeunderstandingworkslarge, chiang2025audioawarelargelanguagemodels, jiang2025s2sarenaevaluatingspeech2speechprotocols} which are capable of processing raw audio for evaluation. However, ALM judges are expensive, opaque, and often still struggle to reason about non-linguistic cues as we show in Sec.~\ref{sec:results}.

\noindent \textbf{TRACE.} To address the  drawbacks of LLM and ALM judges in S2S evaluation, we introduce TRACE (\textbf{T}extual \textbf{R}easoning over \textbf{A}udio \textbf{C}ues for \textbf{E}valuation), a two stage training-free framework that provides auditable and cost-efficient speech evaluation. \textit{Stage 1} compiles a textual blueprint of inexpensive audio signals. \textit{Stage 2} provides an LLM with the blueprint to produce dimension-wise ratings which are fused into an overall judgment.

\noindent \textbf{Existing Benchmark Pitfalls.} To validate the effectiveness of TRACE, we rely on publicly available S2S human preference datasets, namely \textsc{SpeakBench} \cite{manakul2025audiojudgeunderstandingworkslarge} and \textsc{S2S-Arena} \cite{jiang2025s2sarenaevaluatingspeech2speechprotocols}. However, these datasets only provide an overall pairwise rating, in contrast to longstanding established speech labeling protocols that advocate for separate perceptual scales \cite{itut_p835_2003, itut_p808_2021}. These datasets also adopt either no-tie or untyped tie protocols, each of which induces its own failure modes:
\begin{itemize}[wide, nosep, topsep=0pt, labelindent=3pt]
    \item \emph{No-tie} protocols force a winner when both candidate responses are poor.
    \item \emph{Untyped tie} protocols (tie-allowed) do not distinguish whether both candidate responses are acceptable or unacceptable.

\end{itemize}
We show in Sec.~\ref{sec:method} that these artifacts make the original dataset labels \emph{hackable} by transcript‑only evaluators and underweight delivery.

\noindent \textbf{A Standards Aligned Labeling Protocol.}
We instead adopt a labeling protocol that follows the spirit of ITU‑T P.835 (separate perceptual scales) and P.800 subjective methods \cite{itut_p800_1996, itut_p835_2003}. Concretely, our \emph{Human Chain‑of‑Thought (HCoT)} protocol elicits \emph{dimension‑first} pairwise judgments for Content (C), Voice Quality (VQ), and Paralinguistics (P) and an overall label with a typed-tie labeling scheme (\textit{both\_good}, \textit{both\_bad}, or a winner).

\noindent \textbf{Why C, VQ, P?}
Foundational communication research \cite{crystal1975paralinguistics} suggests that all spoken expression can be studied through linguistics and non-linguistics. We map linguistics to Content (C), and we break down non-linguistics into Voice Quality (VQ) and Paralinguistics (P) since there is extensive prior research on VQ (e.g., \cite{lo2019mosnet, Ravuri2023UncertaintyAA, huimosuncertainty}). The C/VQ/P demarcation is also used in benchmarks such as VocalBench \cite{liu2025vocalbenchbenchmarkingvocalconversational} and AudioJudge \cite{manakul2025audiojudgeunderstandingworkslarge}, and each is operationalizable with inexpensive signals at scale.

\noindent \textbf{Contributions.} We summarize our contributions as follows:
\begin{enumerate}
    \item \textbf{HCoT}: A dimension-first, typed-tie re-annotation of \textsc{SpeakBench} and \textsc{S2S-Arena}, yielding reliable, diagnostic labels aligned with ITU-T guidance.
    \item \textbf{TRACE}: A training-free, two-stage evaluator that unlocks the reasoning power of text-LLMs for speech evaluation.
    \item \textbf{Evidence}: Compared to LLM and ALM baselines, TRACE attains higher agreement with HCoT overall ratings and achieves strong fidelity on Paralinguistics, while being significantly cheaper than ALMs.
    \item \textbf{Release}: We release the HCoT re-annotations and TRACE framework to support reproducibility and future work.
\end{enumerate}

\begin{figure*}[t]
\centering

\includegraphics[height=0.245\linewidth]{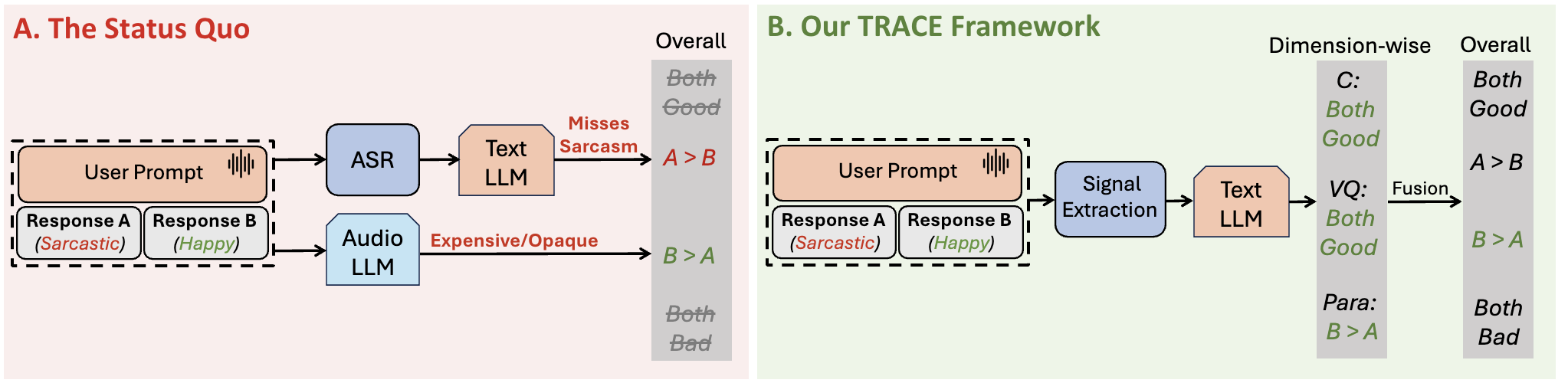}
\caption{\textbf{Bridging the Evaluation Gap.} \textbf{(A) The Status Quo:} Current transcript-only LLM judges are blind to paralinguistics whereas ALM judges are opaque and expensive, and benchmarks often force winners on noisy data. \textbf{(B) TRACE Framework:} We introduce HCoT, a dimension-first speech labeling protocol, and TRACE. TRACE extracts acoustic features into a structured blueprint, allowing text-LLMs to reason over audio for S2S evaluation.}
\label{fig:main}
\end{figure*}

\section{Related Work}

\noindent \textbf{S2S Models.} Recent advances in multimodal learning have enabled speech-to-speech voice assistants \cite{openai2024gpt4ocard, comanici2025gemini25pushingfrontier, zhang2023speechgptempoweringlargelanguage, xie2024miniomnilanguagemodelshear, fang2025llamaomniseamlessspeechinteraction, défossez2024moshispeechtextfoundationmodel}. A simple approach to S2S models is to build a cascade system, wrapping an LLM with automatic speech recognition (ASR) to transcribe spoken input and text-to-speech (TTS) to produce spoken output \cite{chen2024voicebenchbenchmarkingllmbasedvoice}. While straightforward to implement, cascade systems suffer from excessive latency \cite{li2025predgenacceleratedinferencelarge}, error propagation from ASR \cite{min2025endtoendoverkillrethinkingcascaded}, and limited access to paralinguistic information \cite{jiang2025s2sarenaevaluatingspeech2speechprotocols}.

To address these limitations, recent models such as LLaMA-Omni and Moshi \cite{fang2025llamaomniseamlessspeechinteraction, défossez2024moshispeechtextfoundationmodel} operate end-to-end, directly modeling speech input and speech output through discrete speech tokens or embeddings \cite{zhan2025anygptunifiedmultimodalllm, zeng2024glm4voiceintelligenthumanlikeendtoend, wang2024freezeomnismartlowlatency}.

\noindent \textbf{S2S Benchmarks.}
The rapid progress in S2S modeling has motivated new benchmarks. Early efforts such as VoiceBench \cite{chen2024voicebenchbenchmarkingllmbasedvoice} assess general knowledge of S2S models using response transcripts, while subsequent works \cite{liu2025vocalbenchbenchmarkingvocalconversational, hou2025sovabenchbenchmarkingspeechconversation} incorporate objective acoustic metrics such as word error rate (WER) and emotion score. A complementary line of work emphasizes human preference judgments. These benchmarks \cite{jiang2025s2sarenaevaluatingspeech2speechprotocols, manakul2025audiojudgeunderstandingworkslarge, chiang2025audioawarelargelanguagemodels} collect pairwise or point-wise human ratings, providing valuable testbeds for automatic evaluation methods that align with human judgment. However, these datasets collapse multiple perceptual dimensions into a single overall score, reducing reliability and diagnostic value. We address this gap by re-annotating existing benchmarks with dimension-wise ratings and extending preference modeling \citep{rao1967ties,davidson1970ties} to include typed-ties. 

%In contrast, our work introduces a training-free evaluator that explicitly scores (C, VQ, EQ), providing dimension-wise judgments that are interpretable and yield stronger agreement with human annotations than vanilla LLM/ALMs. %and ALM-based approaches.

\noindent \textbf{S2S Auto-Raters.}
Several approaches have been proposed for automatically evaluating S2S models. One strategy is to use ASR to transcribe spoken output and then prompt an LLM judge, yielding a training-free baseline that is easy to scale but blind to acoustic cues \cite{chen2024voicebenchbenchmarkingllmbasedvoice, liu2025vocalbenchbenchmarkingvocalconversational, hou2025sovabenchbenchmarkingspeechconversation}. Another approach employs ALMs that analyze audio directly and produce judgments \cite{chiang2025audioawarelargelanguagemodels, manakul2025audiojudgeunderstandingworkslarge, jiang2025s2sarenaevaluatingspeech2speechprotocols}. A third line of work trains dedicated evaluators via instruction tuning or reinforcement learning from preference data \cite{ji2025wavrewardspokendialoguemodels,ge2025sagelmmultiaspectexplainablelarge}, which can better align with human ratings but may not generalize to new domains. 

\noindent \textbf{Acoustic Metrics.} Modern non-intrusive predictors estimate perceptual speech quality without a reference \cite{reddy2021dnsmos, mittag2021nisqa, lo2019mosnet}. Prior works from affective computing also provide feature sets for expressive speech: 
the openSMILE toolkit \citep{eyben2010opensmile}, 
the eGeMAPS minimalistic acoustic parameter set \citep{eyben2016egemaps}, 
and the ComParE challenge series \citep{schuller2013compare}. Our acoustic blueprints compile and reuse these features.

\noindent \textbf{Text-to-Text Evaluation.}
There are a numerous related works that use LLM judges to evaluate the output of text-to-text models \cite{unieval, geval, branchsolvemerge, dnaeval, healthbench}. Some of these efforts draw parallels with our work, pointing to the importance of decomposing an evaluation metric into multiple dimensions \cite{dnaeval} and taking a rubric-and-aggregation approach \cite{branchsolvemerge}. However, these methods are constrained to textual inputs, and they do not consider the unique challenges of \textit{speech}-to-\textit{speech} evaluation such as tone, emphasis, intonation, rhythm, accent, voiced emotion, etc., all of which are integral to speech interaction. We bridge this gap by using lightweight, off-the-shelf tools to extract audio primitives, which are given to an LLM for dimension-wise evaluation. This structured, multi-tool approach is designed to address the unique challenges of speech evaluation.
\section{Method} \label{sec:method}

Our objective is to develop an S2S auto-rater that is efficient, human–centric, and accurate. 
We proceed in two steps: (i) establish and align a multi-aspect benchmark via Human Chain-of-Thought (HCoT) re-annotation; 
(ii) evaluate TRACE against this benchmark. 

\begin{figure*}[t]
\centering

\includegraphics[height=0.35\linewidth]{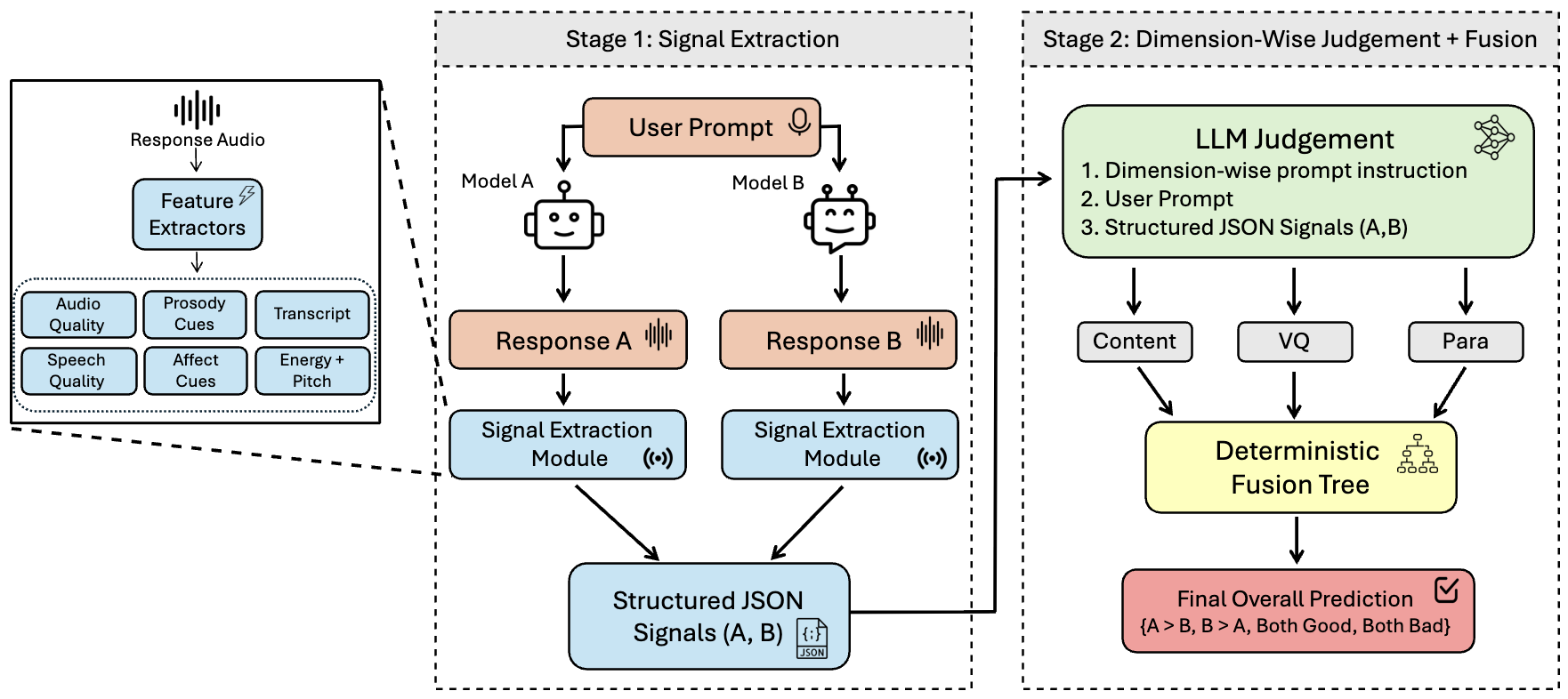}
\caption{\textbf{The TRACE Architecture.} 
\textbf{Phase 1 (Signal Extraction):} We extract inexpensive signals for Content (ASR), Voice Quality (MOS predictors), and Paralinguistics (prosody, affect, energy etc.). 
\textbf{Phase 2 (Inference):} These signals form a structured textual blueprint of audio cues, which is then passed to an LLM judge to make dimension-wise decisions. The dimension-wise deicisons are fused via a deterministic tree to yield the final score.}
\label{fig:blueprint}
\end{figure*}

\subsection{Existing Benchmark Pitfalls}

\begin{table}[t]
  \centering
  \begin{tabular}{lcc}
    \toprule
    & \multicolumn{2}{c}{\textbf{Accuracy (\%)}} \\
    \cmidrule(lr){2-3}
    \textbf{Judge Type} & \textbf{SpeakBench} & \textbf{S2S-Arena} \\
    \midrule
    Random Guess  & 33.3 & 50.0 \\
    Audio Judge   & 51.4 {\scriptsize (47.6–56.6)} & 78.6 {\scriptsize (75.0–81.9)} \\
    LLM Judge     & 59.8 {\scriptsize (55.8–64.6)} & 78.8 {\scriptsize (74.7–82.4)} \\
    \bottomrule
  \end{tabular}
  \caption{\label{tab:llm_beats_audio}
  \textbf{Original labels are "hackable".} Using Gemini 2.5 Flash as the backbone, a text-only LLM Judge exceeds Audio Judge on \textsc{SpeakBench} and \textsc{S2S-Arena} original labels, highlighting that the original labels overweight content. We therefore do \emph{not} adopt the original labels for benchmarking.}
  \vspace{-6pt}
\end{table}

% \afterpage{
% \begin{figure*}[t]
% \centering
% \includegraphics[width=.5\textwidth]{latex/figs/speakbench_bar_chart.pdf}\hfill
% \includegraphics[width=.5\textwidth]{latex/figs/s2s_bar_chart.pdf}
% \caption{\textbf{Accuracy comparison across judges}. Per-dimension accuracy on SpeakBench (left) and S2S-Arena (right). JSON-Judge enhances perceptual \emph{audio} dimensions (VQ/EQ) without sacrificing Content accuracy, resulting in stronger alignment with human judgments under the HCoT labeling scheme.}
% \label{fig:newsroom}
% \end{figure*}
% }

% \afterpage{
% \begin{table*}[t]
%   \centering\small
%   \begin{tabular}{lccc}
%   \toprule
%   \textbf{Cost Category} & \textbf{AudioJudge} & \textbf{LLM Judge} & \textbf{JSON-Judge} \\
%   \midrule
%   GPU Cost (\$)  & 0.000 & 0.256 & 0.424 \\
%   API Cost (\$)  & 12.532 & 2.507 & 3.734 \\
%   \midrule
%   \textbf{Total (\$)}  & \textbf{12.532} & \textbf{2.763} & \textbf{4.158} \\
%   \bottomrule
%   \end{tabular}
%   \caption{\label{tab:judge_cost_analysis}\textbf{Cost on \textsc{SpeakBench} with GPT-4o}. JSON-Judge is $\sim$3$\times$ cheaper than AudioJudge. Full breakdown in Appx.~\ref{app:cost_analysis}.}
% \end{table*}
% }

We study the \textit{pairwise} S2S evaluation setting using \textsc{SpeakBench} \cite{manakul2025audiojudgeunderstandingworkslarge} and the English subset of \textsc{S2S-Arena} \cite{jiang2025s2sarenaevaluatingspeech2speechprotocols}. \textsc{SpeakBench} consists of spoken instructions calling for certain types of content, paralinguistic features, and vocal styles, while \textsc{S2S-Arena} contains explicit instructions, perception-probing tasks, and real-life scenarios.
Original overall-only labels in both benchmarks exhibit two issues:

\noindent \textbf{(1) Reliability.}
Protocols can misrepresent user experience: \textsc{S2S-Arena} forces a winner even when both responses are unacceptable; \textsc{SpeakBench} permits untyped ties that fail to distinguish whether both responses are acceptable or unacceptable. On our \textsc{S2S-Arena} subset, we find that paralinguistics is rated \emph{both-bad} in a majority of examples ($\approx$55\%, see Appendix Tab.~\ref{tab:annotations}), which explains why forced-winner protocols fabricate superiority and motivates our typed ties (both-good, both-bad).

\noindent \textbf{(2) Multi-aspect validity.}
Humans judge speech via separable dimensions (C, VQ, P). 
Under the original labels, a transcript-only LLM judge can match or outperform a more expensive ALM judge (Tab.~\ref{tab:llm_beats_audio}), indicating that existing benchmark labels overweight textual content and underweight audio/paralinguistic factors; this undermines certification of truly human-centric \emph{speech} evaluators.

\subsection{HCoT Protocol}

To address these limitations, we introduce a Human Chain-of-Thought (HCoT) annotation protocol. 

Formally, a user audio input $P$ elicits two candidate audio responses $(A,B)$ from different S2S systems. Under the HCoT protocol, expert raters first produce \emph{dimension-first} (DF) pairwise judgments for Content (C), Voice Quality (VQ), and Paralinguistics (P) followed by an an overall label $\hat{Y}$. The label space for the dimension-wise and overall ratings indicates which response is preferred or whether the pair forms a \emph{typed tie}: $\{1,2,\text{both-good},\text{both-bad}\}$. A typed tie discriminates a tie amongst \{both-bad, both-good\}. This yields (i) an auditable path from parts to whole and (ii) a clean separation of \emph{acceptability} (pass/fail) from \emph{superiority} (which acceptable one is better):

\[
f(P,A,B)\;\rightarrow\;\bigl\{\hat{Y},\,\Delta_C,\,\Delta_{VQ},\,\Delta_{P}\bigr\},
\]

with $\Delta_{(\bullet)}\in\{1,2,\text{both-good},\text{both-bad}\}$. 

It is important to note that \textit{typed ties} introduce a hybrid absolute/relative grading scheme since the \textit{both-bad} threshold is absolute whereas picking a winner is relative. To decouple these effects, a rating is assigned (by a human or automatic judge) in the following manner:
\begin{enumerate}
\item Decide if $A$ is acceptable along the dimension evaluated (C, VQ, P, or Overall).
\item Decide if $B$ is acceptable along the dimension evaluated (C, VQ, P, or Overall).
\item If exactly one is acceptable, return $1$ or $2$.
\item If both are unacceptable, return both-bad.
\item If both are acceptable, do a relative comparison and return $1$, $2$, or both-good.
\end{enumerate}

This hybrid $\textrm{absolute} \to \textrm{relative}$ rating scheme allows us to separate acceptable plateaus (\textit{both-good}) from unacceptable holes (\textit{both-bad}), enabling diagnosis and meaningful reporting of ties.

\noindent \textbf{Standards alignment.}
Our dimension-first labeling mirrors ITU--T guidance: P.835 advocates \emph{separate} ratings for the speech signal, background noise, and overall quality; P.800 enumerates core subjective procedures (ACR/CCR/DCR); and P.808 specifies crowdsourced protocols with quality control \cite{itut_p800_1996, itut_p835_2003, itut_p808_2021}.
This provides an external rationale for typed ties and for decoupling \emph{acceptability} (both-bad/both-good) from \emph{superiority} (winner).%\citep{itut_p835_2003,itut_p800_1996,itut_p808_2021}

\begin{table*}[ht]
\centering\small
\begin{tabular}{lccc}
\toprule
\multicolumn{4}{c}{\textsc{SpeakBench}}\\
\midrule
Pair & 2-way & 3-way & 4-way \\
\midrule
orig $\leftrightarrow$ blind & \textbf{92.8} {\scriptsize (89.0-95.8)} (N{=}237) & \textbf{67.1} {\scriptsize (62.9-71.2)} (N{=}496) & - \\
orig $\leftrightarrow$ HCoT & \textbf{94.2} {\scriptsize (90.9-96.7)} (N{=}274) & \textbf{70.1} {\scriptsize (65.9-74.1)} (N{=}495) & - \\
blind $\leftrightarrow$ HCoT & \textbf{97.5} {\scriptsize (95.0-98.9)} (N{=}278) & \textbf{76.3} {\scriptsize (72.5-80.0)} (N{=}494) & \textbf{74.9} {\scriptsize (70.9-78.5)} (N{=}494) \\
\midrule
\multicolumn{4}{c}{\textsc{S2S-Arena}}\\
\midrule
Pair & 2-way & 3-way & 4-way \\
\midrule
orig $\leftrightarrow$ blind & \textbf{89.4} {\scriptsize (82.9-94.3)} (N{=}123) & - & - \\
orig $\leftrightarrow$ HCoT & \textbf{92.9} {\scriptsize (86.7-96.5)} (N{=}113) & - & - \\
blind $\leftrightarrow$ HCoT &\ \textbf{99.0} {\scriptsize (94.1-100.0)} (N{=}101) & \textbf{88.9} {\scriptsize (85.0-92.0)} (N{=}314) & \textbf{87.6} {\scriptsize (83.4-90.8)} (N{=}314) \\
\bottomrule
\end{tabular}
\vspace{-3pt}
\caption{\label{tab:human-consistency}\textbf{Human-human agreement} (\%) [95\% CI] between label sets on overall decisions. 
2-way: winners only (ties dropped); 3-way: \{1,2,tie\} (typed ties collapsed); 
4-way: \{1,2,\texttt{both\_good},\texttt{both\_bad}\}.}
\end{table*}

\subsection{The TRACE Auto-Rater}

\noindent \textbf{Architecture.}
\emph{TRACE} is a two-stage, training-free evaluator. Stage~1 extracts inexpensive audio signals for each dimension and assembles them into a compact blueprint. Stage~2 prompts an LLM with the blueprints to produce dimension-wise judgments and reasoning for each rating (no overall score is requested). A deterministic fusion rule then maps $(\Delta_C,\Delta_{VQ},\Delta_{P})$ to the overall label $\hat{Y}$. 

\noindent \textbf{Stage 1: Evidence blueprint.}
\begin{lstlisting}[style=jsonstyle,caption={Truncated Exemplar}]
{
  "A": {"asr_text": "...", 
        "mos_overall": 3.9,
        "prosody": {"pitch": 142, "rate": 155}, "affect": {"calm": 0.62}},
  "B": {"asr_text": "...", 
        "mos_overall": 4.2,
        "prosody": {"pitch": 128, "rate": 162}, "affect": {"calm": 0.35}}
}
\end{lstlisting}
For each response, we construct a structured feature set capturing:
\emph{Content} (ASR transcript); 
\emph{Voice Quality} (objective speech-quality indicators); 
\emph{Paralinguistics} (prosodic descriptors, affect/intent style cues, and simple accent proxies). For \emph{Content}, we use the Whisper-large-v3 model \cite{radford2022whisper}. For \emph{Voice Quality} we rely on non-intrusive speech quality predictors (e.g., DNSMOS~P.835), and for \emph{Paralinguistics} we use lightweight prosody/affect descriptors \citep{mittag2021nisqa,reddy2021dnsmos,eyben2010opensmile,eyben2016egemaps, ma2023emotion2vecselfsupervisedpretrainingspeech}.

\noindent \textbf{Stage 2: LLM judgment (Dimension-First).}
\begin{lstlisting}[
  caption={Per-Dimension Decisions (no overall requested).},
  label={lst:llm-output}
]
{
  "prediction_content": "1",
  "prediction_vq": "both_good",
  "prediction_para": "2",
  "reasoning": { "content": "...", "vq": "...", "para": "..." }
}
\end{lstlisting}
The LLM receives the user prompt and the two candidate response  blueprints. It is instructed to output a structured JSON with \emph{per-dimension} decisions. This forces reasoning over distilled, human-aligned signals rather than raw audio, improving interpretability and stability while enabling the LLM to reason over audio cues that are crucial for speech evaluation.

\noindent \textbf{Zero-shot fusion (algorithms in Appendix).}
\emph{TRACE} is training-free and uses a deterministic fusion policy with one dataset-specific \emph{policy prior} to reflect benchmark intent, similar to previous works \cite{lee-etal-2025-checkeval}. For \textsc{SpeakBench} (instruction-following), our fusion rule prioritizes content-first and uses non-content dimensions as tie-breakers to mirror the original dataset intent. For \textsc{S2S-Arena} (perception and delivery centric), many responses suffer from paralinguistic failures that, in the context typical of this dataset, render them completely unacceptable. Therefore, we apply an \emph{acceptability cap} that forces the predicted overall rating to be $\preceq \Delta_{P}$ and $\Delta_C$ so that such responses are not overrated. Full pseudocode appears in App.~\ref{app:fusion}.

\subsection{Evaluation}

\noindent \textbf{Human–human reference.} We report the inter–human agreement for overall labels to contextualize automatic judges and annotation strategies (Tab.~\ref{tab:human-consistency}).

\noindent \textbf{Metrics and uncertainty.}
Our primary endpoint is 4-way \emph{accuracy} over $\{1,2,\text{both-good},\text{both-bad}\}$:
$$
\text{Acc} \;=\; \frac{1}{N}\sum_{i=1}^{N}\mathbf{1}\{\hat{Y}_i = Y_i.\}
$$
We report 95\% confidence intervals for overall label accuracy via \emph{paired} non-parametric bootstrap, preserving example-wise pairing across systems.

\begin{table*}[t]
  \centering\small
  \begin{tabular}{llcccc}
    \toprule
    \textbf{Dataset} & \textbf{Judge} & \textbf{\;\;\; Content \;\;\;} & \textbf{Voice Quality} & \textbf{Paralinguistics} & \textbf{Overall} \\
    \midrule
    \multirow{5}{*}{\textsc{SpeakBench}}
      & Random Guess   & 25.0 & 25.0 & 25.0 & 25.0 \\
      & Audio Judge    & 62.5 & 45.6 & 21.4 & 61.1 {\scriptsize (56.7–65.4)} \\
      & LLM Judge      & 60.4 & 39.8 & 29.8 & 62.7 {\scriptsize (58.2–67.0)} \\
      & \textbf{TRACE} & \textbf{63.2} & \textbf{50.4} & \textbf{39.6} & \textbf{68.6} {\scriptsize (64.3–72.7)} \\
      & Human–human agreement & 76.0 & 60.0 & 82.0 & 60.0 \\
    \midrule
    \multirow{5}{*}{\textsc{S2S-Arena}}
      & Random Guess   & 25.0 & 25.0 & 25.0 & 25.0 \\
      & Audio Judge    & \textbf{58.9} & 52.5 & 37.7 & 47.5 {\scriptsize (42.4–52.7)} \\
      & LLM Judge      & 57.0 & \textbf{55.1} & 35.4 & 45.9 {\scriptsize (40.4–51.3)} \\
      & \textbf{TRACE} & 58.0 & 51.6 & \textbf{48.1} & \textbf{57.0} {\scriptsize (51.6–62.4)} \\
      & Human–human agreement & 73.3 & 48.3 & 75.0 & 75.0\\
    \bottomrule
  \end{tabular}
  \vspace{-3pt}
  \caption{\label{tab:main-acc}\textbf{Per-dimension and overall accuracy vs.\ HCoT}. Using Gemini 2.5 Flash as the backbone, TRACE improves VQ/P while maintaining Content parity on \textsc{SpeakBench}, and yields the largest gains on Paralinguistics for \textsc{S2S-Arena}.}
\end{table*}

\section{Results} \label{sec:results}

We organize the results as follows: (i) validate the HCoT benchmark; (ii) report top-line accuracies; (iii) show \emph{how} TRACE uses delivery cues (mechanism) and \emph{why} it succeeds; (iv) quantify cost-efficiency; (v) assess robustness (backbone) and results relative to human-human agreement. Unless otherwise specified, all experiments use Gemini 2.5 Flash \cite{comanici2025gemini25pushingfrontier} as the backbone model.

\subsection{Validation of HCoT Annotations}
\label{sec:validation}

\noindent \textbf{Annotation Protocol.} Each dataset was rated in three passes. First, a \emph{blind overall-first} rating captured the original human impression without any dimensional guidance. 
Second, a full \emph{HCoT} (Human Chain-of-Thought) re-annotation collected dimension-first judgments for \{C, VQ, P\} and then overall. Finally, we \emph{randomly resampled} a subset from each dataset and performed a second independent HCoT rating to test repeatability and record inter-human agreement.

\noindent \textbf{Coherence.} We verify that a simple multinomial logistic model using human $(C,VQ,P)$ reconstructs the HCoT overall label at high accuracy on both datasets, indicating overall is a low-noise function of \{C,VQ,P\}. Typed ties and dataset-intent policies (content-first on \textsc{SpeakBench}; acceptability cap on \textsc{S2S-Arena}) make acceptability explicit and eliminate forced-winner artifacts. 

\noindent\textbf{Reliability across datasets (Cohen's $\kappa$).}
We quantify inter-label agreement using Cohen's chance-corrected coefficient
$\kappa=\frac{p_o-p_e}{1-p_e}$, where $p_o$ is observed agreement and $p_e$ is
the chance agreement from rater marginals \citep{cohen1960}.
On \textsc{SpeakBench}, blind$\leftrightarrow$HCoT agreement on the typed 4-way overall
label $\{1,2,\textit{both\_good},\textit{both\_bad}\}$ is
$\kappa=0.651$ ($N{=}468$; 95\%~CI $[0.596,0.702]$).
On \textsc{S2S-Arena}, blind$\leftrightarrow$HCoT (4-way) is $\kappa=0.796$
($N{=}314$; 95\%~CI $[0.740,0.849]$). 

\noindent
\emph{Interpretation of $\kappa$:} values 0.61–0.80 denote
\textit{substantial} and 0.81–1.00 \textit{almost perfect} agreement
\citep{landis1977}.
%; per-dimension human bands are omitted in main but consistent with the overall range (appendix).

\subsection{Overall and per-dimension performance}
\label{sec:perf}
Tab.~\ref{tab:main-acc} summarizes accuracy against the HCoT labels. TRACE improves VQ/P while maintaining Content parity on \textsc{SpeakBench}, and yields the largest gains on Paralinguistics for \textsc{S2S-Arena}. We assessed paired differences between judges using two-sided McNemar tests \cite{mcnemar1947} on item-wise correctness (vs.\ HCoT overall labels). For \textsc{S2S-Arena}, TRACE outperforms both the audio-only and transcript-only judges ($p{<}10^{-3}$ for all comparisons), and remains significantly higher than the LLM judge ($p{=}0.0017$). On \textsc{SpeakBench}, TRACE also exceeds the LLM judge ($p{=}0.02$) and the Audio Judge ($p{<}10^{-3}$). These tests confirm that observed gains are statistically reliable.

As an ablation, we also try applying the majority voting fusion rule from \cite{manakul2025audiojudgeunderstandingworkslarge} as a fusion policy, in which a majority vote is taken amongst $\{\Delta_C, \Delta_{VQ}, \Delta_{P}\}$. Results are reported in Appendix  Tab.~\ref{tab:voting_vs_tree_fusion}. We observe that our acoustic blueprint and our tree-based fusion rule have additive benefits. Under either fusion rule, TRACE outperforms its Audio-Judge and LLM-Judge counterparts, and when applied on top of any judge our tree-fusion rule outperforms its majority voting counterpart.

\noindent \textbf{\emph{What} TRACE does and \emph{why} it succeeds?}
We probe \emph{how} judges use audio cues with two probes (same fusion rules as \S\ref{sec:method}): (P1) a content-controlled counterfactual (force \texttt{Content=both\_good} and record which dimension (if any) resolves the tie and its decision accuracy); (P2) one-at-a-time ablations (flip a single dimension to \texttt{both\_good} and report overall flip rate). We also analyze (P3) judge performance when a legitimate winner exists and when one does not. We use $Y^\star$ to denote the ground-truth overall label:

\smallskip
\noindent\textbf{(P1) Content-controlled Counterfactual.}
Set $\tilde C=\texttt{both\_good}$ and compute $\tilde Y=f(\tilde C,VQ,P)$.
We record which dimension \emph{resolves the tie}
(\,$P$ if $P\in\{1,2\}$ before $VQ$, else $VQ$ if $VQ\in\{1,2\}$\,), or \emph{tie} if
$\tilde Y=\texttt{both\_good}$. 

% We also report \emph{decision accuracy}
% $\mathbbm{1}[\tilde Y=Y^\star]$ conditioned on the resolving dimension.

\smallskip
\noindent\textbf{(P2) One-at-a-time Ablations.}
For each $D\in\{C,VQ,P\}$, replace $D\!\leftarrow\!\texttt{both\_good}$ and recompute
$Y_D=f(C',VQ',P')$. The \emph{flip rate} for $D$ is
$\Pr\!\big[Y_D\neq f(C,VQ,P)\big]$, quantifying dependence on that dimension.

\smallskip
\noindent
\textbf{(P3) Attributing Performance.} We compute from the fused overall prediction $\hat Y$ (using the same fusion policy as in the main evaluation) and the HCoT overall label $Y^\star \in \{1,2,\texttt{both\_good},\texttt{both\_bad}\}$:
\[
\underbrace{\Pr\!\big(\hat Y\in\{1,2\}\mid Y^\star=\texttt{both\_bad}\big)}_{\text{\emph{Winner-on-bad} (lower is better)}}
\]
\[
\underbrace{\Pr\!\big(\hat Y=Y^\star\mid Y^\star\in\{1,2\}\big)}_{\text{\emph{Winner-slice accuracy} (higher is better)}}.
\]
Winner-on-bad measures fabricated winners on unacceptable pairs; winner-slice accuracy measures correctness when a legitimate winner exists. Together they explain \emph{why} a judge’s overall improves or degrades and complement P1–P2’s mechanism-focused probes.

\begin{figure}[t]
\centering
\includegraphics[width=.98\columnwidth]{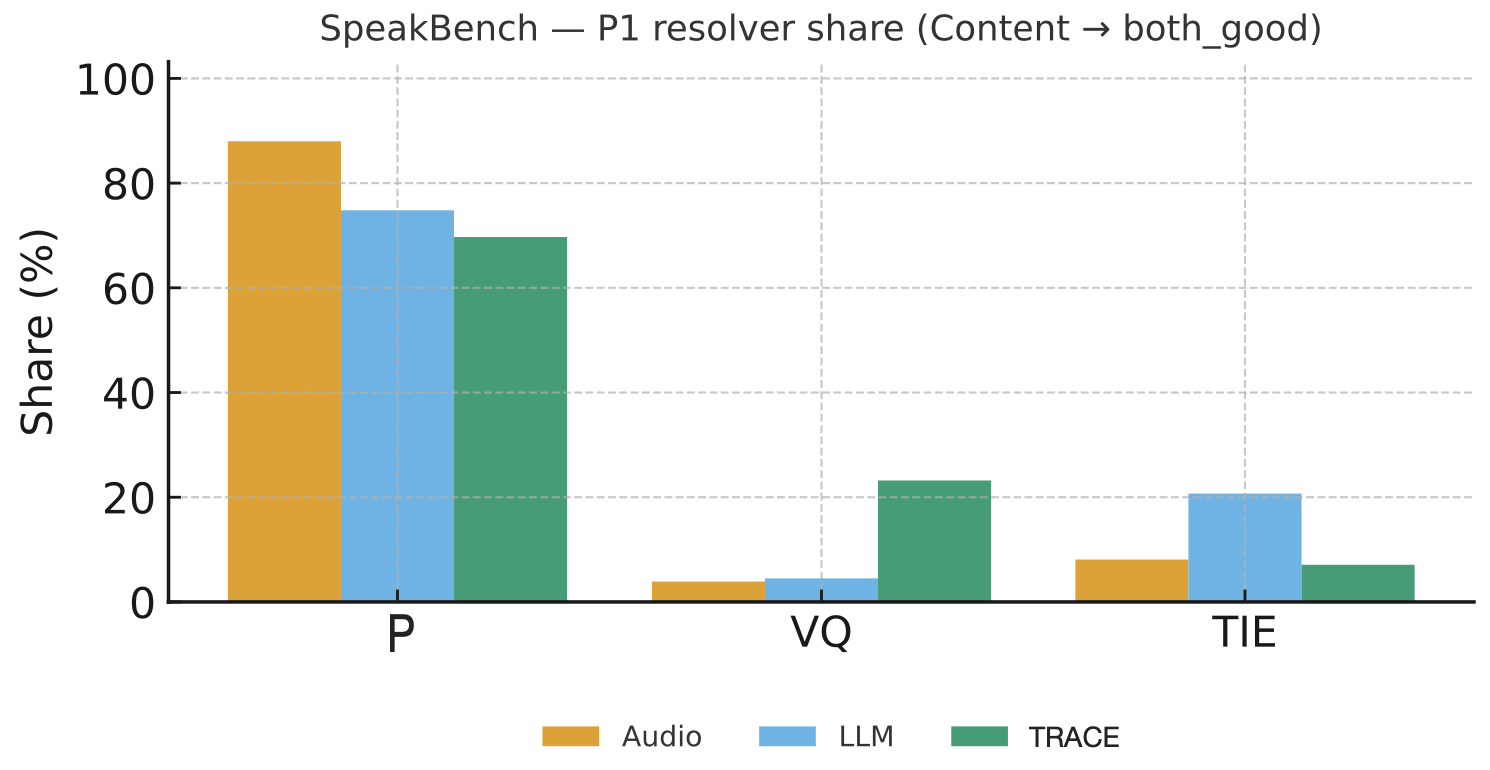}
\vspace{-5pt}
\caption{\textbf{P1 Counterfactual (\textsc{SpeakBench}).} TRACE \emph{selectively} uses delivery (VQ) to break semantic ties (VQ share $\approx$23\% vs.\ $\sim$3-5\% for baselines).}
\label{fig:behave_mech}
\end{figure}
\begin{figure}[t]
\centering
\includegraphics[width=.98\columnwidth]{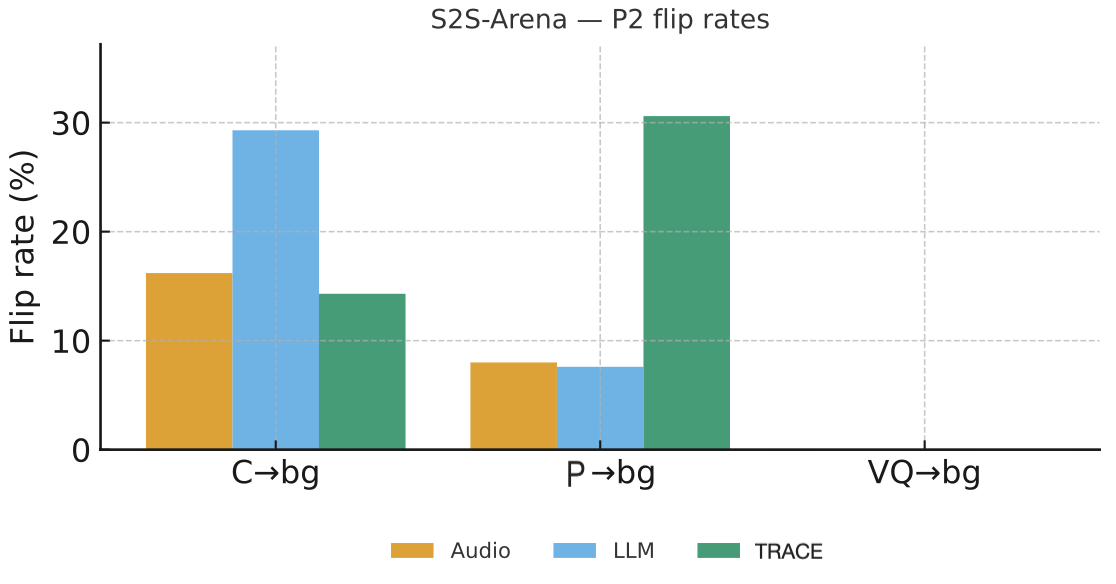}
\vspace{-5pt}
\caption{\textbf{P2 Flip Rates (\textsc{S2S-Arena})}. TRACE is significantly more sensitive to Paralinguistics than LLM-Judge or Audio Judge.}
\label{fig:behave_flip}
\end{figure}
\begin{figure}[t]
\centering
\includegraphics[width=.98\columnwidth]{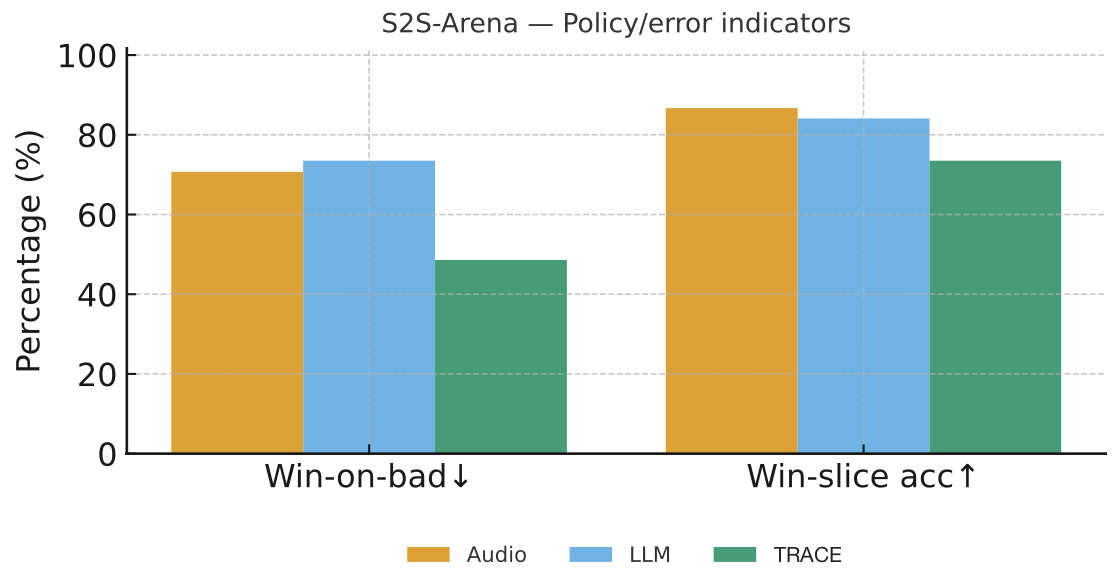}
\vspace{-5pt}
\caption{\textbf{P3 Attributing Performance (\textsc{S2S-Arena}).} With many \texttt{both-bad} pairs (58\%), TRACE cuts \emph{winner-on-bad} from $\sim$70–74\% to \textbf{48.6\%}.}
\label{fig:behave_outcome}
\vspace{-4pt}
\end{figure}

\noindent\textbf{Takeaways.} (i) \emph{Selective delivery use:} on \textsc{SpeakBench}, TRACE leans on VQ/P to resolve content ties far more often (Fig.~\ref{fig:behave_mech}). 
(ii) Fig.~\ref{fig:behave_flip} reveals that TRACE is significantly more sensitive to Paralinguistics compared to other judges, namely, modifying Paralinguistics results in changing the overall decision. 
(iii) \emph{Policy-aware fusion:} on \textsc{S2S-Arena}, where \texttt{both\_bad} pairs are common, we see fabricated winners are suppressed, driving the overall gain (Fig.~\ref{fig:behave_outcome}). TRACE reduces \emph{winner-on-bad} to \textbf{48.6\%} (Audio: 70.7\%, LLM: 73.5\%), which largely explains its overall win (Tab.~\ref{tab:main-acc}). The trade-off is that, on the \emph{winner slice}, TRACE’s accuracy is \textbf{73.5\%} (Audio: 86.7\%, LLM: 84.1\%), indicating headroom when good winners exist. 

\begin{figure*}[ht]
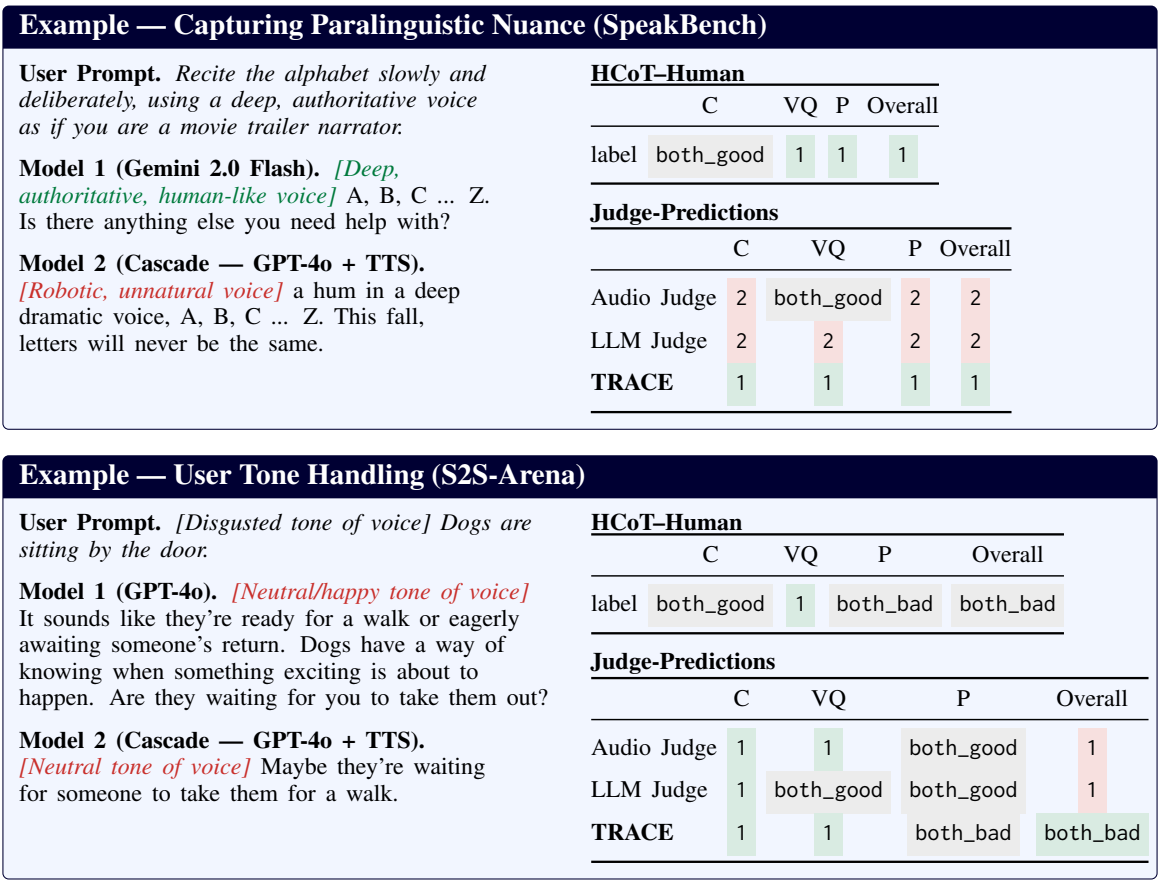

\centering

% ================= EXAMPLE 1 =================
\begin{tcolorbox}[
  colback=softblue,
  colframe=blue!20!black,
  fonttitle=\bfseries,
  title=Example — Capturing Paralinguistic Nuance (SpeakBench),
  boxrule=0.4pt,
  arc=2pt,
  left=3pt,right=3pt,top=3pt,bottom=3pt,
  width=0.95\textwidth,
  halign=center
]
\setlength{\parskip}{3pt}

\raggedright{
\begin{minipage}[t]{0.42\linewidth}\raggedright\small
\setlength{\parskip}{1pt}

\textbf{User Prompt.}  
\emph{Recite the alphabet slowly and deliberately, using a deep, authoritative voice as if you are a movie trailer narrator.}\\

\medskip
\textbf{Model 1 (Gemini 2.0 Flash).}  
\emph{\textcolor{goodgreen}{[Deep, authoritative, human-like voice]}}  
A, B, C ... Z. Is there anything else you need help with?\\

\medskip
\textbf{Model 2 (Cascade — GPT-4o + TTS).}  
\emph{\textcolor{badred}{[Robotic, unnatural voice]}}  
a hum in a deep dramatic voice, A, B, C ... Z. This fall, letters will never be the same.

\end{minipage}}\hspace{34pt}
\begin{minipage}[t]{0.42\linewidth}\small
\setlength{\tabcolsep}{2pt}

\textbf{HCoT--Human}  
\begin{tabular}{@{}lcccc@{}}
\toprule
& C & VQ & P & Overall\\
\midrule
label & \lblcol{gray}{both\_good} & \lblcol{goodgreen}{1} & \lblcol{goodgreen}{1} & \lblcol{goodgreen}{1}\\
\bottomrule
\end{tabular}

\medskip
\textbf{Judge-Predictions}  
\begin{tabular}{@{}lcccc@{}}
\toprule
& C & VQ & P & Overall \\
\midrule
Audio Judge & \lblcol{badred}{2} & \lblcol{gray}{both\_good} & \lblcol{badred}{2} & \lblcol{badred}{2} \\
LLM Judge & \lblcol{badred}{2} & \lblcol{badred}{2} & \lblcol{badred}{2} & \lblcol{badred}{2} \\
\textbf{TRACE} & \lblcol{goodgreen}{1} & \lblcol{goodgreen}{1} & \lblcol{goodgreen}{1} & \lblcol{goodgreen}{1} \\
\bottomrule
\end{tabular}

\end{minipage}
\end{tcolorbox}

%\vspace{4pt}
\vspace{-4pt}

% ================= EXAMPLE 2 =================
\begin{tcolorbox}[
  colback=softblue,
  colframe=blue!20!black,
  fonttitle=\bfseries,
  title=Example — User Tone Handling (S2S-Arena),
  boxrule=0.4pt,
  arc=2pt,
  left=3pt,right=3pt,top=3pt,bottom=3pt,
  width=0.95\textwidth,
  halign=center
]
\setlength{\parskip}{3pt}

\begin{minipage}[t]{0.49\linewidth}\raggedright\small
\setlength{\parskip}{4pt}

\textbf{User Prompt.}  
\emph{[Disgusted tone of voice] Dogs are sitting by the door.}\\

\medskip
\textbf{Model 1 (GPT-4o).}  
\emph{\textcolor{badred}{[Neutral/happy tone of voice]}}\\
It sounds like they're ready for a walk or eagerly awaiting someone's return. Dogs have a way of knowing when something exciting is about to\\ happen. Are they waiting for you to take them out?\\

\medskip
\textbf{Model 2 (Cascade — GPT-4o + TTS).}\\  
\emph{\textcolor{badred}{[Neutral tone of voice]}}  
Maybe they're waiting\\ for someone to take them for a walk.

\end{minipage}\hfill
%\begin{minipage}[t]{0.42\linewidth}\raggedright\small
\begin{minipage}[t]{0.49\linewidth}\raggedright\small
\setlength{\tabcolsep}{2pt}

\textbf{HCoT--Human}  
\begin{tabular}{@{}lcccc@{}}
\toprule
& C & VQ & P & Overall\\
\midrule
label & \lblcol{gray}{both\_good} & \lblcol{goodgreen}{1} & \lblcol{gray}{both\_bad} & \lblcol{gray}{both\_bad}\\
\bottomrule
\end{tabular}

\medskip
\textbf{Judge-Predictions}  
\begin{tabular}{@{}lcccc@{}}
\toprule
& C & VQ & P & Overall \\
\midrule
Audio Judge & \lblcol{goodgreen}{1} & \lblcol{goodgreen}{1} & \lblcol{gray}{both\_good} & \lblcol{badred}{1} \\
LLM Judge & \lblcol{goodgreen}{1} & \lblcol{gray}{both\_good} & \lblcol{gray}{both\_good} & \lblcol{badred}{1} \\
\textbf{TRACE} & \lblcol{goodgreen}{1} & \lblcol{goodgreen}{1} & \lblcol{gray}{both\_bad} & \lblcol{goodgreen}{both\_bad} \\
\bottomrule
\end{tabular}

\end{minipage}
\end{tcolorbox}

%\vspace{2pt} 
\vspace{-5pt}
\caption{TRACE mirrors human preference by leveraging VQ/P cues to break content ties \textbf{(top)}, and correctly flags emotionally inappropriate responses \textbf{(bottom)}. Green = agree w/ human, red = disagree.}
\label{fig:example_speakbench_eq_vq}
\end{figure*}

\noindent \textbf{Efficiency and Scalability.}
TRACE operates on inexpensive structured signals; Stage~1 features are batchable and Stage~2 passes textual JSON objects to an LLM. On \textsc{SpeakBench} (GPT-4o), TRACE is $\sim$3$\times$ cheaper than Audio Judge while achieving higher overall accuracy (Tab.~\ref{tab:judge_cost_analysis}). See Appendix Tab.~\ref{tab:full_judge_cost_analysis} for a detailed breakdown of the cost analysis.

% === Keep the cost table in main ===
\begin{table}[h]
  \centering\small
  \begin{tabular}{lccc}
  \toprule
  \textbf{Cost} & \textbf{Audio Judge} & \textbf{LLM Judge} & \textbf{TRACE} \\
  \midrule
  GPU (\$)  & 0.00 & 0.25 & 0.42 \\
  API (\$)  & 12.53 & 2.51 & 3.73 \\
  \midrule
  \textbf{Total (\$)}  & \textbf{12.532} & \textbf{2.763} & \textbf{4.158} \\
  \bottomrule
  \end{tabular}
  \caption{\label{tab:judge_cost_analysis}\textbf{Cost on \textsc{SpeakBench} with GPT-4o}. TRACE is $\sim$3$\times$ cheaper than AudioJudge while remaining more accurate.}
\end{table}

\noindent \textbf{Robustness (backbone).}
Replacing Gemini 2.5 Flash with GPT-4o preserves trends: TRACE wins on non-content dimensions and overall (Tab.~\ref{tab:judge_model_ablation}).

\begin{table}[h]
  \centering\small
  \begin{tabular}{lcccc}
    \toprule
    \textbf{Judge} & \textbf{Content} & \textbf{VQ} & \textbf{Para} & \textbf{Overall} \\
    \midrule
     Audio Judge & 51.4 & 39.1 & 17.2 & 53.4 \\
      LLM Judge  & \textbf{58.8} & 32.1 & 29.8 & 60.6 \\
      \textbf{TRACE} & 58.0 & \textbf{50.0} & \textbf{36.2} & \textbf{62.1} \\
    \bottomrule
  \end{tabular}
  \caption{\label{tab:judge_model_ablation}\textbf{Backbone ablation with GPT-4o}. TRACE gains are robust across backbones (\textsc{SpeakBench}).}
\end{table}

% \begin{table}[h]
%   \centering\small
%   \begin{tabular}{llcccc}
%     \toprule
%     \textbf{Dataset} & \textbf{Backbone} & \textbf{Content} & \textbf{VQ} & \textbf{P} & \textbf{Overall} \\
%     \midrule
%     \multirow{3}{*}{\textsc{SpeakBench} (HCoT)}
%       & AudioJudge (GPT-4o) & 51.4 & 39.1 & 17.2 & 53.4 \\
%       & LLM Judge (GPT-4o)  & \textbf{58.8} & 32.1 & 29.8 & 60.6 \\
%       & \textbf{JSON-Judge (GPT-4o)} & 58.0 & \textbf{50.0} & \textbf{36.2} & \textbf{62.1} \\
%     \bottomrule
%   \end{tabular}
%   \caption{\label{tab:judge_model_ablation}\textbf{Backbone ablation}. JSON-Judge gains are robust across backbones.}
% \end{table}

% \begin{table*}[h]
%   \centering\small
%   \begin{tabular}{llcccc}
%     \toprule
%     \textbf{Dataset} & \textbf{Backbone} & \textbf{Content} & \textbf{VQ} & \textbf{P} & \textbf{Overall} \\
%     \midrule
%     \multirow{3}{*}{\textsc{SpeakBench} (HCoT)}
%       & AudioJudge (GPT-4o) & 51.4 & 39.1 & 17.2 & 53.4 \\
%       & LLM Judge (GPT-4o)  & \textbf{58.8} & 32.1 & 29.8 & 60.6 \\
%       & \textbf{JSON-Judge (GPT-4o)} & 58.0 & \textbf{50.0} & \textbf{36.2} & \textbf{62.1} \\
%     \bottomrule
%   \end{tabular}
%   \caption{\label{tab:judge_model_ablation}\textbf{Backbone ablation}. JSON-Judge gains are robust across backbones.}
% \end{table*}

%meow
%\input{latex/human_human_agreement_table}

\noindent \textbf{Human-human agreement.} Relative to the human–human upper bound (Tab.~\ref{tab:main-acc}), TRACE approaches the lower edge on \textsc{SpeakBench} and closest to lower band on \textsc{S2S-Arena} in comparison to other judge methods. The residual gap concentrates where delivery is the sole differentiator or rater criteria are stricter.

\noindent \textbf{Qualitative case studies.}
A representative prompt from \textsc{SpeakBench} and \textsc{S2S-Arena} (content ties; delivery differs) shows TRACE is the only automatic judge to mirror human preference (Fig.~\ref{fig:example_speakbench_eq_vq}).

%, and upstream detector brittleness.
% \begin{small}
% \begin{tabular}{@{}p{\dimexpr\linewidth-12pt\relax}@{}} % \linewidth is now width inside tcolorbox
% \toprule
% \textbf{Example: Capturing Paralinguistic Nuances (SpeakBench)} \\
% \midrule
% \textbf{User Prompt:} \\
% \textit{Recite the alphabet slowly and deliberately, using a deep, authoritative voice as if you are a movie trailer narrator.} \\
% \midrule
% \textbf{Model 1 (Gemini 2.0 Flash):} \\
% \textit{"[Deep, authoritative, human-like voice] Okay, I will do that for you. A, B, C...Z. Is there anything else you need help with?"} \\
% \midrule
% \textbf{Model 2 (Cascade - GPT-4o + TTS):} \\
% \textit{"[Robotic, unnatural voice] a hum in a deep dramatic voice, A, B, C...Z. This fall, letters will never be the same."} \\
% \midrule
% \textbf{HCoT Human Labels:} \\
% Content: Both Good, \ VQ: 1, \ EQ: 1, \ Overall: 1 \\
% \midrule
% \textbf{Judge Predictions:} \\
% \begin{tabular}{@{}l c c c c@{}}
% \textbf{Judge} & \textbf{Content} & \textbf{VQ} & \textbf{P} & \textbf{Overall} \\
% \midrule
% Audio Judge & 2 & both\_good & 2 & 2 \\
% LLM Judge & 2 & 2 & 2 & 2 \\
% \textbf{JSON-Judge} & 1 & \textbf{1} & \textbf{1} & \textbf{1} \\
% \bottomrule
% \end{tabular} \\
% \bottomrule
% \end{tabular}
% \end{small}
% %\end{tcolorbox}
% \caption{An example from SpeakBench where JSON-Judge correctly captures the nuances of Voice Quality (VQ) and Paralinguistics (P). Model 1 fulfills the stylistic prompt, while Model 2's robotic delivery fails. Only JSON-Judge mirrors the human preference.}
% \label{fig:example_speakbench_eq_vq}
% \end{figure}

\section{Conclusion}

We introduced HCoT and TRACE, a structured, multi-aspect framework for evaluating speech-to-speech systems that captures content, voice quality, and paralinguistic attributes. Our experiments show that TRACE more accurately predicts human judgments across multiple dimensions, outperforming approaches that rely on raw audio or transcripts alone. TRACE is scalable, interpretable, and flexible, enabling fine-grained evaluation of speech systems. Future work includes extending the framework to handle multilingual scenarios, richer prosodic features, and real-time evaluation, further bridging the gap between automatic judges and human perception.

%\begin{center}
\paragraph{Limitations:}
%\end{center}
While TRACE provides a structured, interpretable alternative to direct audio or text-based evaluation, several limitations remain. First, our experiments are limited to English datasets (\textsc{SpeakBench} and \textsc{S2S-Arena}); the generality of the framework across languages and cultural norms of expressivity remains to be tested. Second, the current acoustic schema was designed manually. Although it captures core perceptual dimensions (content, voice quality, paralinguistics), it may omit finer-grained attributes that enable it to handle edge cases. Future work could explore data-driven schema induction that adapt the feature extraction stage dynamically to new tasks. Finally, TRACE relies on upstream automatic extractors whose errors can propagate into the final judgment. Addressing this dependency through calibration or confidence weighting/optimization
is a promising future direction.

\paragraph{Ethical considerations:} Our work pertains to automatic evaluation of speech-to-speech voice assistants. While there is great potential for voice assistants to do at lot of good in the world (e.g. accesibility, healthcare, therapy), there is also potential for them to do harm. This could happen intentionally when voice assistants are used for malicious purposes (e.g. fraud, harassment, misinformation), or unintentionally when a flawed voice assistant has harmful failure modes (e.g. giving bad therapeutic advice). This makes the ethics of automated judges complex, as a flawed judge might foster overconfidence in flawed voice assistants, while a strong judge could accelerate the development of strong voice assistants that could be used for malicious purposes. It is important that these issues are discussed and addressed both inside and outside of the research community.

\section{Acknowledgments}

This research was supported by an Amazon Gift Award and the National Science Foundation grant CPS-2317079.

\bibliography{custom}
\clearpage

\section{Appendix}

\appendix

\renewcommand{\thesubsection}{\thesection.\arabic{subsection}} % subsections as A.1, A.2, ...

\section{Datasets}  

\subsection{SpeakBench Dataset}
The SpeakBench dataset \cite{manakul2025audiojudgeunderstandingworkslarge} originally contains $N=508$ examples. Each example is a tuple $(P, A, B)$ where $P$ is a user prompt, $A$ is response candidate A, and $B$ is response candidate B. Each example also has a human label indicating either a winner or an untyped tie, i.e., one of $\{A, B, \text{tie}\}$.  
We filter out examples that were used for few-shot prompting in the original paper, leaving $N=497$ examples. 

\subsection{S2S-Arena Dataset}
The S2S-Arena dataset \cite{jiang2025s2sarenaevaluatingspeech2speechprotocols} originally contains $N=457$ examples. We restrict ourselves to the English subset, which contains $N=314$ examples. Each example is a tuple $(P, A, B)$ where $P$ is a user prompt, $A$ is response candidate A, and $B$ is response candidate B. Each example has a human label indicating a winner, i.e., either $A$ or $B$.

\section{HCoT Re-Annotation}

\paragraph{Annotation process.} We annotate both the SpeakBench and S2S-Arena dataset using our proposed HCoT annotation protocol. Specifically, we introduce dimension-wise annotations along Content, Voice Quality, and Paralinguistics, followed by an overall rating. The overall and dimension-wise labels either declare a winner among the two response candidates, or they indicate both responses are good, or both responses are bad (typed ties). Two annotators, both native English speakers, independently annotated one of the two full datasets following these guidelines. The instructions provided to the annotators are the same as the instructions provided to the judge models (Tab.~\ref{tab:system_prompt_hcot}), and the annotation results are summarized in Tab.~\ref{tab:annotations}.

\paragraph{Annotators.} The two annotators are students who are native English speakers. Recruitment was informal, and as student workers they were compensated for annotating via the same standard stipend process as all other student labor. Both annotators were made aware of how their ratings would be used and consented to their use in this work and their public release.

\begin{table*}[t]
  \centering\small
  \begin{tabular}{llccc}
    \toprule
    \textbf{Dataset} & \textbf{Label} & \textbf{Both Bad} & \textbf{Winner} & \textbf{Both Good} \\
    \midrule
    \multirow{4}{*}{\textsc{SpeakBench}}
      & Content & 86 & 293 & 118 \\
      & Voice Quality & 71 & 282 & 144 \\
      & Paralinguistics & 383 & 99 & 15 \\
      & Overall & 85 & 366 & 46 \\
    \midrule
    \multirow{4}{*}{\textsc{S2S-Arena}}
      & Content & 80 & 161 & 73 \\
      & Voice Quality & 14 & 138 & 162 \\
      & Paralinguistics & 173 & 91 & 50 \\
      & Overall & 181 & 113 & 20 \\
    \bottomrule
  \end{tabular}
  \caption{\label{tab:annotations}\textbf{HCoT Annotation Counts.} High prevalence of Paralinguistics \textit{both-bad} in \textsc{S2S-Arena} motivates typed ties and the acceptability cap.}
\end{table*}

\newpage

\section{Structured JSON Feature Vector}  

\subsection{JSON Schema}

Stage 1 of our proposed TRACE evaluator builds a JSON feature vector for each audio response. The schema of the JSON is shown in Fig.~\ref{fig:json_schema}.

\subsection{Features \& Model Specifications}

The JSON used to represent each audio response contains several fields derived from both open-source models and basic audio processing libraries. The features and their sources are described below:

\paragraph{agent\_response:}
A transcription of the audio response generated by the \texttt{openai/whisper-large-v3} model (Hugging Face).

\paragraph{agent\_emotion:}
A vector of emotion scores computed from the \texttt{iic/emotion2vec\_plus\_large} model (Hugging Face).

\paragraph{agent\_accent:}
A vector of accent cosine similarity scores from the \texttt{Jzuluaga/accent-id-commonaccent\_ecapa} model (Hugging Face).

\paragraph{agent\_audio\_quality:}
A set of audio quality scores generated by DNSMOS and P.808-based models from \citet{reddy2021dnsmos}. It includes:
\begin{itemize}
    \item \textit{DNSMOS\_Personalized\_Signal\_Quality}: Signal quality from DNSMOS (1–5).  
    \item \textit{DNSMOS\_Personalized\_Background\_Quality}: Background noise quality from DNSMOS (1–5).  
    \item \textit{DNSMOS\_Personalized\_Overall\_Quality}: Overall naturalness and audio quality from DNSMOS (1–5).  
    \item \textit{P808\_Overall\_Quality}: Overall audio quality following ITU-T P.808 recommendation (1–5).  
\end{itemize}

\paragraph{agent\_audio\_properties:}
Low-level acoustic properties of the audio response extracted using signal processing or audio analysis tools:
\begin{itemize}
    \item \textit{Mean\_Pitch\_Hz}, \textit{Std\_Dev\_Pitch\_Hz}, \textit{Full\_Pitch\_Contour\_Hz}: Pitch statistics and contour.  
    \item \textit{Integrated\_Loudness\_LUFS}, \textit{Std\_Dev\_Loudness\_LUFS}, \textit{Full\_Loudness\_Contour\_LUFS}: Loudness statistics and contour.  
    \item \textit{Speech\_Rate\_WPM}, \textit{Articulation\_Rate\_WPM}: Speech rate in words per minute, including and excluding pauses.  
\end{itemize}

These features are readily computed using basic Python libraries (e.g., \texttt{librosa}, \texttt{aubio}, and \texttt{pyloudnorm}). They provide a structured representation of both the content and acoustic quality of each agent response, allowing for detailed evaluation along multiple dimensions. We provide an example of the generated JSON file in Fig.~\ref{fig:json_example}.

\begin{figure*}[t]  % [t] for top of page
\centering
\begin{minipage}{0.95\textwidth} % almost full width
\begin{lstlisting}[basicstyle=\ttfamily\small,breaklines=true]
{
  "agent_response": "a transcription of the response",
  "agent_emotion": "a vector of emotion scores for the agent's response from the emotion2vec model",
  "agent_accent": "a vector of cosine similarity scores for the agent's accent",
  "agent_audio_quality": {
    "DNSMOS_Personalized_Signal_Quality": "signal quality score from DNSMOS model (1-5, higher is better)", 
    "DNSMOS_Personalized_Background_Quality": "background noise quality score from DNSMOS model (1-5, higher is better)", 
    "DNSMOS_Personalized_Overall_Quality": "overall naturalness and audio quality score from DNSMOS model (1-5, higher is better)",
    "P808_Overall_Quality": "overall naturalness and audio quality score from P.808 recommendation standard (1-5, higher is better)"
  },
  "agent_audio_properties": {
    "Mean_Pitch_Hz": "mean pitch (fundamental frequency) of agent response",
    "Std_Dev_Pitch_Hz": "standard deviation in pitch",
    "Full_Pitch_Contour_Hz": "full pitch contour",
    "Integrated_Loudness_LUFS": "average loudness of the agent response measured in LUFS",
    "Std_Dev_Loudness_LUFS": "standard deviation in loudness",
    "Full_Loudness_Contour_LUFS": "full loudness contour",
    "Speech_Rate_WPM": "speech rate in words per minute", 
    "Articulation_Rate_WPM": "speech rate in words per minute excluding pauses and gaps in speech"
  }
}
\end{lstlisting}
\end{minipage}
\caption{JSON schema for audio responses.}
\label{fig:json_schema}
\end{figure*}

\clearpage
\begin{figure*}[h!]  % [t] for top of page
\centering
\begin{minipage}{0.95\textwidth} % almost full width
\begin{lstlisting}[basicstyle=\ttfamily\small,breaklines=true]
{
  "agent_response": " Okay, I will do that for you. The sentence is, madam, in Eden, I'm Adam. Now in reverse order it is, Adam, I'm in Eden, madam.",
  "agent_emotion": {
    "angry": 0.0,
    "disgusted": 0.0,
    "fearful": 0.0,
    "happy": 0.0,
    "neutral": 1.0,
    "other": 0.0,
    "sad": 0.0,
    "surprised": 0.0,
    "unknown": 0.0
  },
  "agent_accent": {
    "england": 0.264,
    "us": 0.691,
    "canada": 0.427,
    "australia": 0.313,
    "indian": 0.194,
    "scotland": 0.158,
    "ireland": 0.083,
    "african": 0.164,
    "malaysia": 0.243,
    "newzealand": 0.215,
    "southatlandtic": 0.223,
    "bermuda": 0.152,
    "philippines": 0.126,
    "hongkong": 0.256,
    "wales": 0.182,
    "singapore": 0.127
  },
  "agent_audio_quality": {
    "DNSMOS_Personalized_Signal_Quality": "4.48 / 5.00",
    "DNSMOS_Personalized_Background_Quality": "4.70 / 5.00",
    "DNSMOS_Personalized_Overall_Quality": "4.31 / 5.00",
    "P808_Overall_Quality": "4.20 / 5.00"
  },
  "agent_audio_properties": {
    "Mean_Pitch_Hz": 139.38,
    "Std_Dev_Pitch_Hz": 25.25,
    "Full_Pitch_Contour_Hz": [
      119.94, 130.23, 148.61, 124.17, 91.05, 123.88, 120.5, 131.58, 112.35, 131.03, 
      145.38, 144.2, 185.4, 170.45, 161.96, 161.18, 153.16, 154.28, 174.33, 101.06
    ],
    "Integrated_Loudness_LUFS": -18.78,
    "Std_Dev_Loudness_LUFS": 4.2,
    "Full_Loudness_Contour_LUFS": [
      -23.45, -20.35, -23.09, -28.69, -20.6, -23.38, -18.99, -27.6, -21.36, -26.42, 
      -21.84, -25.28, -16.23, -17.19, -23.11, -10.25, -22.71, -18.09, -18.8, -25.16
    ],
    "Speech_Rate_WPM": 169.57,
    "Articulation_Rate_WPM": 206.35
  }
}
\end{lstlisting}
\end{minipage}
\caption{Example of a JSON feature vector generated for a voice assistant response (\textsc{SpeakBench}).}
\label{fig:json_example}
\end{figure*}

\clearpage
\section{Prompts}
Stage 2 of our proposed TRACE evaluator provides the JSON feature vector for each candidate response along with the user prompt to an LLM. The LLM is prompted to generate dimension-wise labels only. We also replicate this setup with a transcript-only LLM baseline and an Audio LLM Judge. The results of these experiments appear in Tab.~\ref{tab:main-acc} and Tab.~\ref{tab:judge_model_ablation}. The prompts used for these experiments are shown in Tab.~\ref{tab:system_prompt_hcot} and Tab.~\ref{tab:user_prompts_hcot}. 

We also show in Tab.~\ref{tab:llm_beats_audio} that a transcript-only LLM judge matches or beats an Audio LLM Judge on SpeakBench and S2S-Arena using the original dataset labels. The prompts used for this experiment are provided in Tab.~\ref{tab:prompts_llm_beats_audio}.

% Table 1: System Prompt
\begin{table*}[t]
\centering
\small
\begin{tabular}{p{4cm} p{10cm}}
\toprule
\textbf{Prompt Type} & \textbf{Prompt} \\
\midrule
System Prompt (Shared) & You are an evaluator of audio outputs produced by different audio-capable large language models. Your task is to compare two audio
responses (Audio 1 and Audio 2) generated according to a user's instruction. Evaluate based on these criteria: 

1. Content

- Does the content fulfill the user's request accurately?

- Did the content of the response appropriately address the user's instruction?  

2. Voice Quality

- How good is the voice quality of the response?

- Does it sound natural/human, does it mispronounce words, does it have pops or echoes?  

3. Instruction Following Audio:

- Does the response correctly perceive emotion from user's tone of voice, does it correctly express emotion through tone of voice, does it correctly follow paralinguistic instructions?

- This includes both implicit audio instruction like emotional intelligence and explicit audio instruction following.  

Avoid position bias and don't let response length influence your evaluation. After your analysis, output valid JSON with exactly 4 keys:

- ``reasoning'': your explanation of the comparison along each dimension

- ``content'': your rating for content dimension. a string value '1' if the first audio is better, '2' if the second audio is better, 'both\_bad' if they are equally bad, or 'both\_good' if they are equally good

- ``voice\_quality'': your rating for voice quality dimension. a string value '1' if the first audio is better, '2' if the second audio is better, 'both\_bad' if they are equally bad, or 'both\_good' if they are equally good

- ``instruction\_following\_audio'': your rating for instruction following audio dimension. a string value '1' if the first audio is better, '2' if the second audio is better, 'both\_bad' if they are equally bad, or 'both\_good' if they are equally good  

You should only pick a winner along each dimension if there is a clear and obvious difference between the quality of the two responses. If it comes down to minor details, 
then you should opt for using 'both\_bad' or 'both\_good' instead. \\
\bottomrule
\end{tabular}
\caption{Shared System Prompt Across All Judges Used for Judge Comparison on HCoT Labels (Tab.~\ref{tab:main-acc}) and Backbone Ablation (Tab.~\ref{tab:judge_model_ablation})}
\label{tab:system_prompt_hcot}
\end{table*}

% Table 2: User Prompts
\begin{table*}[h!]
\centering
\small
\begin{tabular}{p{4cm} p{10cm}}
\toprule
\textbf{Prompt Type} & \textbf{Prompt} \\
\midrule
User Prompt - Audio Judge & 

Here is the instruction for this test:

\{instruction.wav\}  

Here is the first audio clip: 

\{audio\_a.wav\}  

Here is the second audio clip: 

\{audio\_b.wav\}  

Respond ONLY in text and output valid JSON with keys ``reasoning'', ``content'', ``voice\_quality'', and ``instruction\_following\_audio'':
\\ 
\midrule
User Prompt - LLM Judge & The responses audios (Audio 1 and Audio 2) will be given to you as text transcripts of the response. Since you are only given the transcripts, it is 
okay to make your best guess at rating along each dimension since all of the information needed may not be available.  

Here is the user's input prompt:  

\{user\_prompt\}  

Here is Audio 1 text transcript:  

\{model\_a\_transcript\}  

Here is Audio 2 text transcript:  

\{model\_b\_transcript\}  

Respond ONLY in text and output valid JSON with keys ``reasoning'', ``content'', ``voice\_quality'', and ``instruction\_following\_audio'':
\\ 
\midrule
User Prompt - TRACE & The responses audios (Audio 1 and Audio 2) will be given to you as JSON objects with the following information: 

\{

  ``agent\_response'': a transcription of the response,

  ``agent\_emotion'': a vector of emotion scores for the agent's response from the emotion2vec model,

  ``agent\_accent'': a vector of cosine similarity scores for the agent's accent,

  ``agent\_audio\_quality'':  \{

    ``DNSMOS\_Personalized\_Signal\_Quality'': signal quality score from DNSMOS model (1-5, higher is better),

    ``DNSMOS\_Personalized\_Background\_Quality'': background noise quality score from DNSMOS model (1-5, higher is better),

    ``DNSMOS\_Personalized\_Overall\_Quality'': overall naturalness and audio quality score from DNSMOS model (1-5, higher is better),

    ``P808\_Overall\_Quality'': overall naturalness and audio quality score from P.808 recommendation standard (1-5, higher is better)  
  \},

  ``agent\_audio\_properties'': \{

    ``Mean\_Pitch\_Hz'': mean pitch (fundamental frequency) of agent response,

    ``Std\_Dev\_Pitch\_Hz'': standard deviation in pitch,

    ``Full\_Pitch\_Contour\_Hz'': full pitch contour,

    ``Integrated\_Loudness\_LUFS'': average loudness of the agent response measured in LUFS,  
    ``Std\_Dev\_Loudness\_LUFS'': standard deviation in loudness,

    ``Full\_Loudness\_Contour\_LUFS'': full loudness contour,

    ``Speech\_Rate\_WPM'': speech rate in words per minute,

    ``Articulation\_Rate\_WPM'': speech rate in words per minute excluding pauses and gaps in speech  
    \}

\}

Here is the user's input prompt:  

\{user\_prompt\}  

Here is Audio 1 response JSON:  

\{audio\_a.json\}  

Here is Audio 2 response JSON:  

\{audio\_b.json\}  

Respond ONLY in text and output valid JSON with keys ``reasoning'', ``content'', ``voice\_quality'', and ``instruction\_following\_audio'':
\\ 
\bottomrule
\end{tabular}
\caption{User Prompts for Judge Comparison on HCoT Labels (Tab.~\ref{tab:main-acc}) and Backbone Ablation (Tab.~\ref{tab:judge_model_ablation})}
\label{tab:user_prompts_hcot}
\end{table*}

\begin{table*}[h!]
\centering
\small
\begin{tabular}{p{4cm} p{10cm}}
\toprule
\textbf{Prompt Type} & \textbf{Prompt} \\
\midrule
System Prompt - Audio Judge & You are an evaluator of audio outputs produced by different audio-capable large language models. Your task is to compare two audio
responses (Audio 1 and Audio 2) generated according to a user’s instruction.
Evaluate based on these criteria: 
1. Semantics: Does the content fulfill the user’s request accurately? 
2. Paralinguistics: How well does the speech match requested tone, emotion, style, pacing, and expressiveness?  

Important: Do not favor verbalized descriptions of tone over actual tonal expression. A response that says "I am speaking excitedly" but
sounds flat should rank lower than one that genuinely sounds excited.

Follow this process: 
1. Analyze the key characteristics requested in the user’s instruction 
2. Evaluate how well Audio 1 performs on these characteristics 
3. Evaluate how well Audio 2 performs on these characteristics 
4. Compare their strengths and weaknesses 
5. Decide which is better overall

Avoid position bias and don’t let response length influence your evaluation. After your analysis, output valid JSON with exactly two keys:
’reasoning’ (your explanation of the comparison) and ’label’ (a string value: ’1’ if the first audio is better, ’2’ if the second audio is better, or
’tie’ if they are equally good/bad. Please use "tie" sparingly, and only when you absolutely cannot choose the winner.) \\
\midrule
User Prompt - Audio Judge & Here is the instruction for this test:

\{instruction.wav\}  

Here is the first audio clip: 

\{audio\_a.wav\}  

Here is the second audio clip: 

\{audio\_b.wav\}  

Please analyze which of the two recordings follows the instruction better, or tie. Respond ONLY in text and output valid JSON with keys
’reasoning’ and ’label’ (string, ’1’, ’2’ or ’tie’). \\
\midrule
User Prompt - LLM Judge  & You are an evaluator of audio outputs produced by different audio-capable large language models. Your task is to compare two audio
responses (Audio 1 and Audio 2) generated according to a user’s instruction. Evaluate based on these criteria: 
1. Semantics: Does the content fulfill the user’s request accurately? Did the content of the response appropriately address the user's instruction? 
2. Holistic: How good is the audio response on a holistic user experience level?

Avoid position bias and don’t let response length influence your evaluation. After your analysis, output valid JSON with exactly two keys:
’reasoning’ (your explanation of the comparison) and ’label’ (a string value: ’1’ if the first audio is better, ’2’ if the second audio is better, or
’tie’ if they are equally good/bad.)

The responses audios (Audio 1 and Audio 2) will be given to you as text transcripts of the response. 

Here is the user's input prompt:

\{user\_prompt\}

Here is Audio 1 text transcript: 

\{model\_a\_transcript\}

Here is Audio 2 text transcript: 

\{model\_b\_transcript\}

Respond ONLY in text and output valid JSON with keys ’reasoning’ and ’label’ (string, ’1’, ’2’ or ’tie’). \\
\bottomrule
\end{tabular}
\caption{Prompts used for Audio Judge and LLM Judge on original \textsc{SpeakBench} and \textsc{S2S-Arena} labels (Tab.~\ref{tab:llm_beats_audio}).}
\label{tab:prompts_llm_beats_audio}
\end{table*}

\clearpage
\section{Deterministic Fusion}
\label{app:fusion}

\subsection{Fusion for \textsc{SpeakBench} (content-first, typed ties)}

We use the decision tree shown in Alg.~\ref{alg:speakbench-appx} to fuse the dimension-wise predictions from each judge model to an overall prediction on SpeakBench. The motivation for the decision tree logic (summarized in Tab.~\ref{tab:policy-prior-summary}) is to reflect the original intent of the SpeakBench dataset annotations.

\begin{algorithm}[t]
\caption{SpeakBench Fusion}
\label{alg:speakbench-appx}
\begin{algorithmic}[1]
\Require $\Delta_C,\Delta_{VQ},\Delta_{P}\in\{1,2,\text{both-good},\text{both-bad}\}$
\If{$\Delta_C \in \{1,2\}$} \State \textbf{return} $\hat{Y}\leftarrow \Delta_C$ \EndIf
\If{$\Delta_C = \text{both-good}$}
   \If{$\Delta_{P}\in\{1,2\}$} \State \textbf{return} $\hat{Y}\leftarrow \Delta_{P}$ \EndIf
   \If{$\Delta_{VQ}\in\{1,2\}$} \State \textbf{return} $\hat{Y}\leftarrow \Delta_{VQ}$ \EndIf
   \State \textbf{return} $\hat{Y}\leftarrow \text{both-good}$
\EndIf
\If{$\Delta_C = \text{both-bad}$}
   \If{$\Delta_{P}\in\{1,2\}$} \State \textbf{return} $\hat{Y}\leftarrow \Delta_{P}$ \EndIf
   \If{$\Delta_{VQ}\in\{1,2\}$} \State \textbf{return} $\hat{Y}\leftarrow \Delta_{VQ}$ \EndIf
   \State \textbf{return} $\hat{Y}\leftarrow \text{both-bad}$
\EndIf
\end{algorithmic}
\end{algorithm}

\begin{table*}[h!]
\centering
\small
\begin{tabular}{lcc}
\toprule
\textbf{Dataset} & \textbf{Intent/Priority} & \textbf{One-line Fusion Rule} \\
\midrule
\textsc{SpeakBench} & Delivery $\rightarrow$ Content & Content decides; P/VQ break ties \\
\textsc{S2S-Arena}  & Instruction-following $\rightarrow$ P & \textbf{Cap}: if C or P is both-bad $\Rightarrow$ overall both-bad \\
\bottomrule
\end{tabular}
\caption{\textbf{Policy-prior summary.} Deterministic, monotone fusion with dataset-specific prior.}
\label{tab:policy-prior-summary}
\end{table*}

\subsection{Fusion for \textsc{S2S-Arena} (acceptability cap, typed ties)}
\begin{algorithm}[t]
\caption{S2S-Arena Fusion}
\label{alg:s2sarena-appx}
\begin{algorithmic}[1]
\Require $\Delta_C,\Delta_{VQ},\Delta_{P}\in\{1,2,\text{both-good},\text{both-bad}\}$\\
$\Delta_{\textrm{cap}} \leftarrow \textrm{RatingMin}(\Delta_C, \Delta_{P})$
\If{$\Delta_C \in \{1,2\}$} \State \textbf{return} $\hat{Y}\leftarrow \textrm{RatingMin}(\Delta_C, \Delta_{\textrm{cap}})$ \EndIf
\If{$\Delta_{P}\in\{1,2\}$} \State \textbf{return} $\hat{Y}\leftarrow \textrm{RatingMin}(\Delta_{P}, \Delta_{\textrm{cap}})$ \EndIf
\If{$\Delta_{VQ}\in\{1,2\}$} \State \textbf{return} $\hat{Y}\leftarrow \textrm{RatingMin}(\Delta_{VQ}, \Delta_{\textrm{cap}})$ \EndIf\\
\textbf{return} $\hat{Y}\leftarrow \textrm{RatingMin}(\Delta_C, \Delta_{\textrm{cap}})$
\end{algorithmic}
\end{algorithm}

\begin{algorithm}[t]
\caption{RatingMin operator}
\label{alg:s2sarena-appx-ratingmin}
\begin{algorithmic}[1]
\Require $\Delta_A, \Delta_B\in\{1,2,\text{both-good},\text{both-bad}\}$\\
$\pi_A\leftarrow \begin{cases}
(1,0)\ \textrm{if}\ \Delta_A\!=\!1\\
(0,1)\ \textrm{if}\ \Delta_A\!=\!2\\
(1,1)\ \textrm{if}\ \Delta_A\!=\!\textrm{both-good}\\
(0,0)\ \textrm{if}\ \Delta_A\!=\!\textrm{both-bad}
\end{cases}$\\
$\pi_B\leftarrow \begin{cases}
(1,0)\ \textrm{if}\ \Delta_B\!=\!1\\
(0,1)\ \textrm{if}\ \Delta_B\!=\!2\\
(1,1)\ \textrm{if}\ \Delta_B\!=\!\textrm{both-good}\\
(0,0)\ \textrm{if}\ \Delta_B\!=\!\textrm{both-bad}
\end{cases}$\\
$\pi_C\leftarrow \min(\pi_A, \pi_B)$ \Comment{elementwise min}\\
\textbf{return} $\Delta_C\leftarrow \begin{cases}
1\ \textrm{if}\ \pi_C\!=\!(1,0)\\
2\ \textrm{if}\ \pi_C\!=\!(0,1)\\
\textrm{both-good}\ \textrm{if}\ \pi_C\!=\!(1,1)\\
\textrm{both-bad}\ \textrm{if}\ \pi_C\!=\!(0,0)
\end{cases}$
\end{algorithmic}
\end{algorithm}

\paragraph{Motivation for the acceptability cap.}
In our \textsc{S2S-Arena} slice, Paralinguistics is labeled \textit{both-bad} in 55\% of pairs (on $N{=}314$), 
so forced-winner protocols would fabricate superiority; capping overall by C/P acceptability prevents this artifact. The fusion algorithm is provided in Alg.~\ref{alg:s2sarena-appx}.

\subsection{Comparison to Majority Voting Fusion}

Previous work similar to ours \cite{manakul2025audiojudgeunderstandingworkslarge} introduces a multi-aspect Audio LLM Judge which fuses dimension-wise predictions through a majority vote. We compare this method to our tree-based fusion approach in Tab.~\ref{tab:voting_vs_tree_fusion} and confirm that our tree-based fusion approach unanimously improves overall label accuracy by incorporating dataset intent and policy-aware decision logic.

\section{Sensitivity analysis details}

\paragraph{Procedures.}
We implement three diagnostics to characterize evaluation sensitivity:
(1) \textbf{Content-controlled counterfactual:} force
\texttt{Content=both\_good} and record whether P or VQ determines the
overall label.
(2) \textbf{One-at-a-time ablations:} replace each dimension by
\texttt{both\_good} and measure the fraction of rows where the fused overall
changes.
(3) \textbf{Correctness-aware and policy metrics:} compute
winner-slice accuracy, winner-on-bad rate, and P/VQ decision accuracies
under these perturbations.
All experiments reuse the same fusion functions as the main evaluation:
\textsc{SpeakBench} uses the Content$\rightarrow$P$\rightarrow$VQ tree;
\textsc{S2S-Arena} applies a strict acceptability cap
$\mathrm{overall}\preceq\min(\mathrm{Content},\mathrm{P})$
and, for diagnostic purposes, a lenient ``acceptable'' cap
that permits winners whenever both cues are not \texttt{both\_bad}.

\paragraph{Implementation.}
For each judge we compute the base fused overall label
and the counterfactuals defined above. P/VQ decision accuracy is measured against the ground truth HCoT label whenever a dimension produces a winner in the forced-content counterfactual. Flip rates report the percentage of examples whose overall label changes after
ablation. The same procedure is applied across both datasets.

\begin{table}[t]
  \centering\small
  \begin{tabular}{llc}
    \toprule
    \textbf{Dataset} & \textbf{Judge} & \textbf{Overall} \\
    \midrule
    \multirow{6}{*}{\textsc{SpeakBench}}
      & Audio Judge (voting)     & 58.0 {\scriptsize (53.4–62.3)} \\      
      & Audio Judge (tree)       & 61.1 {\scriptsize (56.7–65.4)} \\
      & LLM Judge (voting)       & 61.2 {\scriptsize (56.5–65.5)} \\
      & LLM Judge (tree)         & 62.7 {\scriptsize (58.2–67.0)} \\
      & TRACE (voting)      & 66.5 {\scriptsize (62.1–70.6)} \\
      & \textbf{TRACE (tree)} & \textbf{68.6} {\scriptsize (64.3–72.7)} \\
    \midrule
    \multirow{6}{*}{\textsc{S2S-Arena}}
      & Audio Judge (voting)     & 37.1 {\scriptsize (32.3–41.9)} \\
      & Audio Judge (tree)       & 47.5 {\scriptsize (42.4–52.7)} \\
      & LLM Judge (voting)       & 34.4 {\scriptsize (29.3–39.8)} \\
      & LLM Judge (tree)         & 45.9 {\scriptsize (40.4–51.3)} \\
      & TRACE (voting)      & 40.4 {\scriptsize (35.0–45.9)} \\
      & \textbf{TRACE (tree)} & \textbf{57.0} {\scriptsize (51.6–62.4)} \\
    \bottomrule
  \end{tabular}
  \caption{\label{tab:voting_vs_tree_fusion}\textbf{Majority Vote Fusion vs.\ Tree-Based Fusion.} Tree-based fusion consistently improves overall accuracy across judges and datasets by incorporating policy-aware decision logic.}
\end{table}

\newpage

\section{Cost Analysis}

We report the total cost of TRACE using GPT-4o on \textsc{SpeakBench} in Tab.~\ref{tab:judge_cost_analysis} of the main text. We provide the full cost breakdown here in Tab.~\ref{tab:full_judge_cost_analysis}. We note that the main cost of Audio Judge is induced by the raw audio input tokens, whereas our method avoids this cost while still capturing speech cues by using textual blueprints of audio signals. Generating these textual blueprints can be done with inexpensive off-the-shelf classifiers and only occurs a small additional GPU cost. Notably, TRACE is $\sim$3$\times$ cheaper than Audio Judge while achieving better performance.

\section{Feature Ablation}

We ablate each of the audio feature groups in TRACE (Tab.~\ref{tab:feature_ablation}) to assess their impact. We find that removing any of the audio features generally degrades performance on Voice Quality or Paralinguistics for at least one of the two benchmarks, motivating their inclusion in the framework. We also note that some variation in performance ($ \pm 3\%$) is likely due to the stochastic output of Gemini 2.5 Flash and should not be over-interpreted. 

\begin{table*}[t]
  \centering\small
  \begin{tabular}{llcccc}
    \toprule
    \textbf{Dataset} & \textbf{Judge} & \textbf{\;\;\; Content \;\;\;} & \textbf{Voice Quality} & \textbf{Paralinguistics} & \textbf{Overall} \\
    \midrule
    \multirow{5}{*}{\textsc{SpeakBench}}
      & Random Guess   & 25.0 & 25.0 & 25.0 & 25.0 \\
      & \textbf{TRACE} & 63.2 & 50.4 & 39.6 & 68.6 {\scriptsize (64.3–72.7)} \\
      & \hspace{3mm}\textit{w/o emotion classifier} & 65.6 & 50.5 & 46.2 & 70.0 {\scriptsize (65.8–74.1)} \\
      & \hspace{3mm}\textit{w/o accent classifier} & 66.3 & 51.9 & 38.9 & 69.2 {\scriptsize (65.0–73.3)} \\
      & \hspace{3mm}\textit{w/o audio quality} & 64.0 & 40.2 & 42.4 & 67.5 {\scriptsize (63.4–71.6)} \\
      & \hspace{3mm}\textit{w/o audio properties} & 64.2 & 48.9 & 38.9 & 67.1 {\scriptsize (62.9–71.2)} \\
    \midrule
    \multirow{5}{*}{\textsc{S2S-Arena}}
      & Random Guess   & 25.0 & 25.0 & 25.0 & 25.0 \\
      & \textbf{TRACE} & 58.0 & 51.6 & 48.1 & 57.0 {\scriptsize (51.6–62.4)} \\
      & \hspace{3mm}\textit{w/o emotion classifier} & 56.4 & 49.4 & 32.8 & 42.4 {\scriptsize (36.9–48.1)} \\
      & \hspace{3mm}\textit{w/o accent classifier} & 57.6 & 50.6 & 46.5 & 58.6 {\scriptsize (53.2–64.0)} \\
      & \hspace{3mm}\textit{w/o audio quality} & 58.3 & 47.5 & 50.6 & 60.2 {\scriptsize (54.8–65.6)} \\
      & \hspace{3mm}\textit{w/o audio properties} & 55.4 & 52.5 & 49.7 & 59.9 {\scriptsize (54.5–65.3)} \\
    \bottomrule
  \end{tabular}
  \vspace{-3pt}
\caption{\label{tab:feature_ablation}
\textbf{TRACE Feature Ablation.} Removing individual audio feature groups generally degrades performance on either Voice Quality or Paralinguistics, although some variation in performance ($ \pm 3\%$) is likely due to the stochastic output of Gemini 2.5 Flash.}

\end{table*}

\begin{table*}[t]
  \centering\small
  \begin{tabular}{lccc}
    \toprule
    \textbf{Cost Category} & \textbf{Audio Judge} & \textbf{LLM Judge} & \textbf{TRACE} \\
    \midrule
    \textbf{Local GPU} & & & \\
    \hspace{2mm}Inference time (hrs) & 0.000 & 0.634 & 1.050 \\
    \hspace{2mm}Rate (\$/hr)         & 0.404 & 0.404 & 0.404 \\
    \hspace{2mm}\textit{Cost (\$)}   & 0.000 & 0.256 & 0.424 \\
    \midrule
    \textbf{API} & & & \\
    \hspace{2mm}Text Input (\$)  & 0.613 & 1.281 & 2.833 \\
    \hspace{2mm}Audio Input (\$) & 10.952 & 0.000 & 0.000 \\
    \hspace{2mm}Text Output (\$) & 0.967 & 1.226 & 0.901 \\
    \hspace{2mm}\textit{Cost (\$)} & 12.532 & 2.507 & 3.734 \\
    \midrule
    \textbf{Total Cost (\$)} & 12.532 & 2.763 & 4.158 \\
    \bottomrule
  \end{tabular}
  \caption{\label{tab:full_judge_cost_analysis}
  \textbf{Detailed Cost analysis on \textsc{SpeakBench}.}
  Total cost of running the evaluation using GPT-4o as the judge backbone.
  Local GPU inference was performed on a single RTX A6000 (48GB), with rental rates estimated from Vast.ai. TRACE is $\sim$3$\times$ cheaper than Audio Judge while achieving better performance.}
\end{table*}

\section{Visual Judge Comparison}

\begin{figure*}[t]
    \centering
    \includegraphics[width=0.85\textwidth]{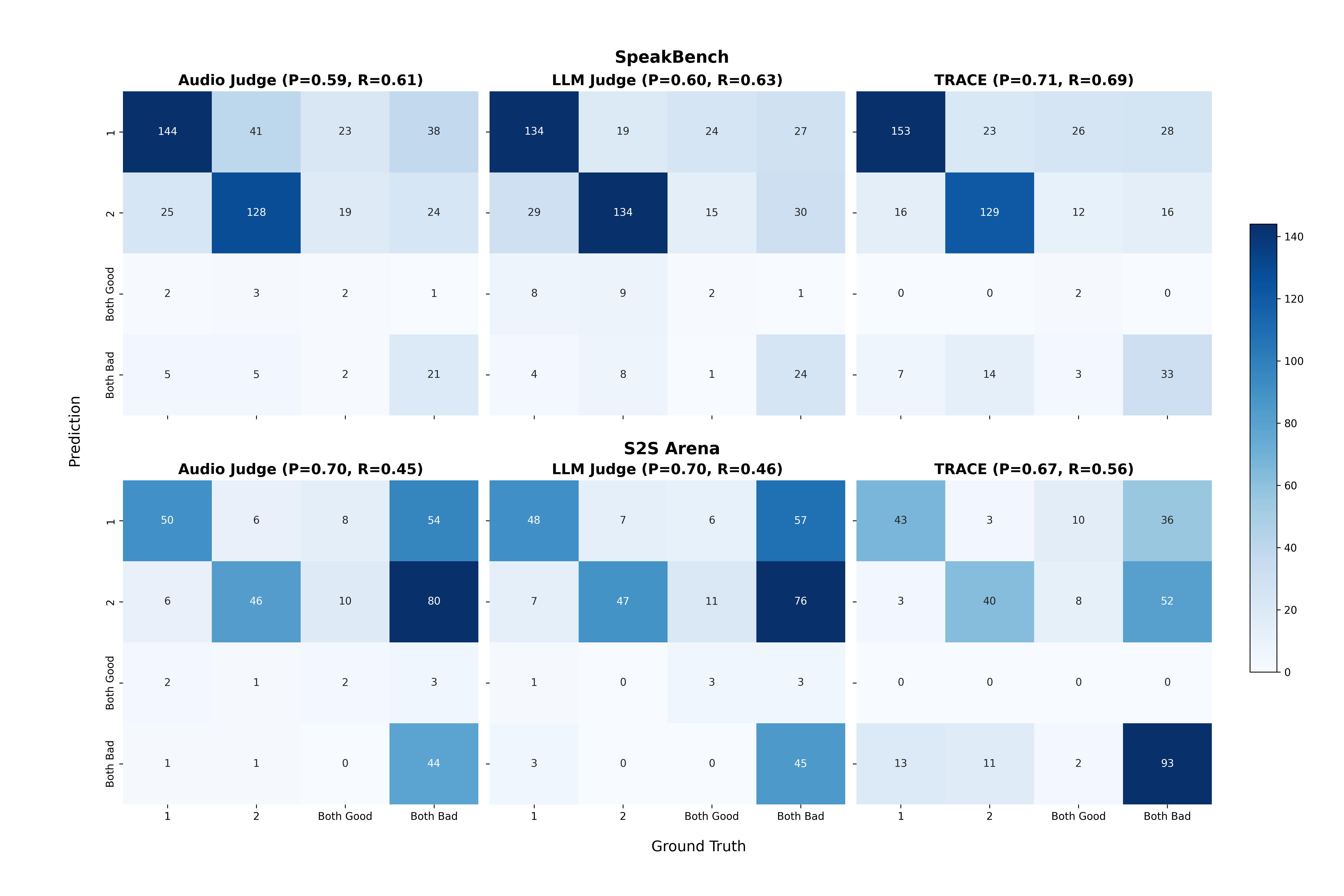}
    \vspace{-10pt}
    \caption{\textbf{Confusion Matrix Comparison.} Fused overall labels predicted by the Audio Judge, LLM Judge, and TRACE against our HCoT overall labels. The underlying model is Gemini 2.5 Flash. “P” and “R” in the plot titles denote precision and recall weighted by class frequency. TRACE achieves the highest recall on both datasets and the highest precision on \textsc{SpeakBench}, indicating stronger alignment with human HCoT annotations.}
    \label{fig:confusion}
\end{figure*}

We include a visual (confusion matrix) comparing judge performance against our HCoT overall labels in Fig.~\ref{fig:confusion}. TRACE achieves the  highest recall on both datasets and the highest precision on \textsc{SpeakBench}, indicating stronger alignment with the HCoT annotations.

\FloatBarrier

\end{document}